\documentclass[lettersize,journal]{IEEEtran}
\usepackage{amsmath,amsfonts}
\usepackage{algorithmic}
\usepackage{algorithm}
\usepackage{array}
\usepackage[caption=false,labelfont=rm,textfont=rm]{subfig}
\usepackage{textcomp}
\usepackage{stfloats}
\usepackage{url}
\usepackage{verbatim}
\usepackage{graphicx}
\usepackage{xcolor}
\def\BibTeX{{\rm B\kern-.05em{\sc i\kern-.025em b}\kern-.08em
    T\kern-.1667em\lower.7ex\hbox{E}\kern-.125emX}}
\usepackage{balance}
\usepackage{multirow}
\usepackage{booktabs} 
\usepackage{hyperref}
\usepackage{bm}
\usepackage{makecell}
\newtheorem{definition}{\bf Definition}[section]

\begin{document}
\title{Automatic Graph Topology-Aware Transformer}
\author{Chao Wang, Jiaxuan Zhao,~\IEEEmembership{Student Member,~IEEE}, Lingling Li,~\IEEEmembership{Senior Member,~IEEE}, Licheng Jiao,~\IEEEmembership{Fellow,~IEEE}, Fang Liu,~\IEEEmembership{Senior Member,~IEEE}, Shuyuan Yang,~\IEEEmembership{Senior Member,~IEEE}

\thanks{This work was supported in part by the Key Scientific Technological Innovation Research Project by Ministry of Education, the State Key Program and the Foundation for Innovative Research Groups of the National Natural Science Foundation of China (61836009), the National Natural Science Foundation of China (U22B2054, U1701267, 62076192, 62006177, 61902298, 61573267, and 62276199), the 111 Project, the Program for Cheung Kong Scholars and Innovative Research Team in University (IRT 15R53), the ST Innovation Project from the Chinese Ministry of Education, the Key Research and Development Program in Shaanxi Province of China(2019ZDLGY03-06), the National Science Basic Research Plan in Shaanxi Province of China(2022JQ-607), the China Postdoctoral fund(2022T150506), the Scientific Research Project of Education Department In Shaanxi Province of China (No.20JY023). (\textit{Corresponding author: Lingling Li}.)}
\thanks{The authors are with the Key Laboratory of Intelligent Perception and Image Understanding of Ministry of Education, International Research Center for Intelligent Perception and Computation, Xidian University, Xi’an 710071, China (e-mail: llli@xidian.edu.cn).}}

\markboth{Journal of \LaTeX\ Class Files,~Vol., No., September 2023}%
{How to Use the IEEEtran \LaTeX \ Templates}

\maketitle

\begin{abstract}
Existing efforts are dedicated to designing many topologies and graph-aware strategies for the graph Transformer, which greatly improve the model's representation capabilities. However, manually determining the suitable Transformer architecture for a specific graph dataset or task requires extensive expert knowledge and laborious trials. This paper proposes an evolutionary graph Transformer architecture search framework (EGTAS) to automate the construction of strong graph Transformers. We build a comprehensive graph Transformer search space with the micro-level and macro-level designs. EGTAS evolves graph Transformer topologies at the macro level and graph-aware strategies at the micro level. Furthermore, a surrogate model based on generic architectural coding is proposed to directly predict the performance of graph Transformers, substantially reducing the evaluation cost of evolutionary search. \textcolor{black}{We demonstrate the efficacy of EGTAS across a range of graph-level and node-level tasks, encompassing both small-scale and large-scale graph datasets.} Experimental results and ablation studies show that EGTAS can construct high-performance architectures that rival state-of-the-art manual and automated baselines \footnote{The code can be accessed at \url{https://github.com/xiaofangxd/EGTAS}.}.
\end{abstract}

\begin{IEEEkeywords}
Graph Transformer, neural architecture search, performance predictor, topology design, graph-aware strategy, graph neural network.
\end{IEEEkeywords}

\section{Introduction}
\IEEEPARstart{G}{raph}-structured data describes a set of concepts (nodes) and their relationships (links), which model many real-world complex problems \cite{8294302,10026148}. Graph neural networks (GNNs) have emerged as a potent tool for extracting high-level node representations from node features and link relationships. This capability enables GNNs to effectively tackle many learning tasks on graph-structured data, such as node classification (NC), graph classification (GC), link prediction (LP), and clustering \cite{JMLR:v23:20-852}. Most GNNs follow the message-passing paradigm that is built by designing aggregation operations to aggregate the information from the neighborhood \cite{9046288,9796468,10255266}. However, as the GNN deepens, the node representations become indistinguishable, which is known as the over-smoothing problem \cite{Li_Han_Wu_2018}.

To alleviate the issue, different topologies are designed by the introduction of skip connections, including ResGCN \cite{Li_2019_ICCV}, JK-net \cite{pmlr-v80-xu18c}, and GCNII \cite{pmlr-v119-chen20v}. Skip connections have the capacity to integrate node features from different levels, thereby augmenting the model capability. However, none of them seems to eliminate the over-smoothing problem completely based on existing experimental observations \cite{10.1145/3485447.3512185}. Moreover, due to its superior ability to extract global information in natural language processing \cite{10.1145/3530811,chao2024match} and computer vision \cite{yang2023video,peng2023usage,MA202366,wu2023video}, the Transformer is introduced into the field of graph learning to enhance the representation ability of the model \cite{muller2023attending,9770130,10092525}. A range of graph-aware strategies are presented to incorporate graph information into vanilla Transformers. Three typical graph-aware strategies include the combination of GNNs and Transformers \cite{NEURIPS2020_94aef384}, the positional embedding with graph information \cite{NEURIPS2019_6e091746}, and the attention matrix with graph information \cite{10.1145/3477495.3532031}. Previous studies \cite{10.1145/3485447.3512185,min2022transformer} demonstrate that the above topologies and graph-aware strategies heavily affect model performance. However, it's important to note that determining the superior architecture and graph-aware strategy often relies heavily on expert knowledge. Consequently, crafting a suitable graph Transformer architecture tailored to a specific graph task or dataset remains a daunting and resource-intensive endeavor.

To reduce the cost of developing GNNs, graph neural architecture search (GNAS) techniques are proposed for automatically designing excellent model architectures \cite{ijcai2021p637}, which provides promising directions to address the above difficulties. A typical GNAS involves exploring a unified search space describing possible architectures, evaluating their performance on downstream tasks, and using optimization algorithms to search for the best architecture \cite{10065594}. GNAS faces two key technical challenges: designing the search space and search strategy. Existing search spaces can be mainly divided into the macro search space and micro search space \cite{wang2022automated}. The former mainly focuses on the design of the GNN topology. The latter focuses on discovering components in GNNs that affect graph information exchange, such as aggregation functions. Recently, Zhang \textit{et al.} \cite{zhang2023autogt} proposed \textcolor{black}{AutoGT} to find optimal graph encoding strategies and model scale of graph Transformers. However, systematic exploration of topologies and graph-aware strategies in graph Transformers is lacking.

\begin{figure*}
\centering
\includegraphics[width=0.7\textwidth]{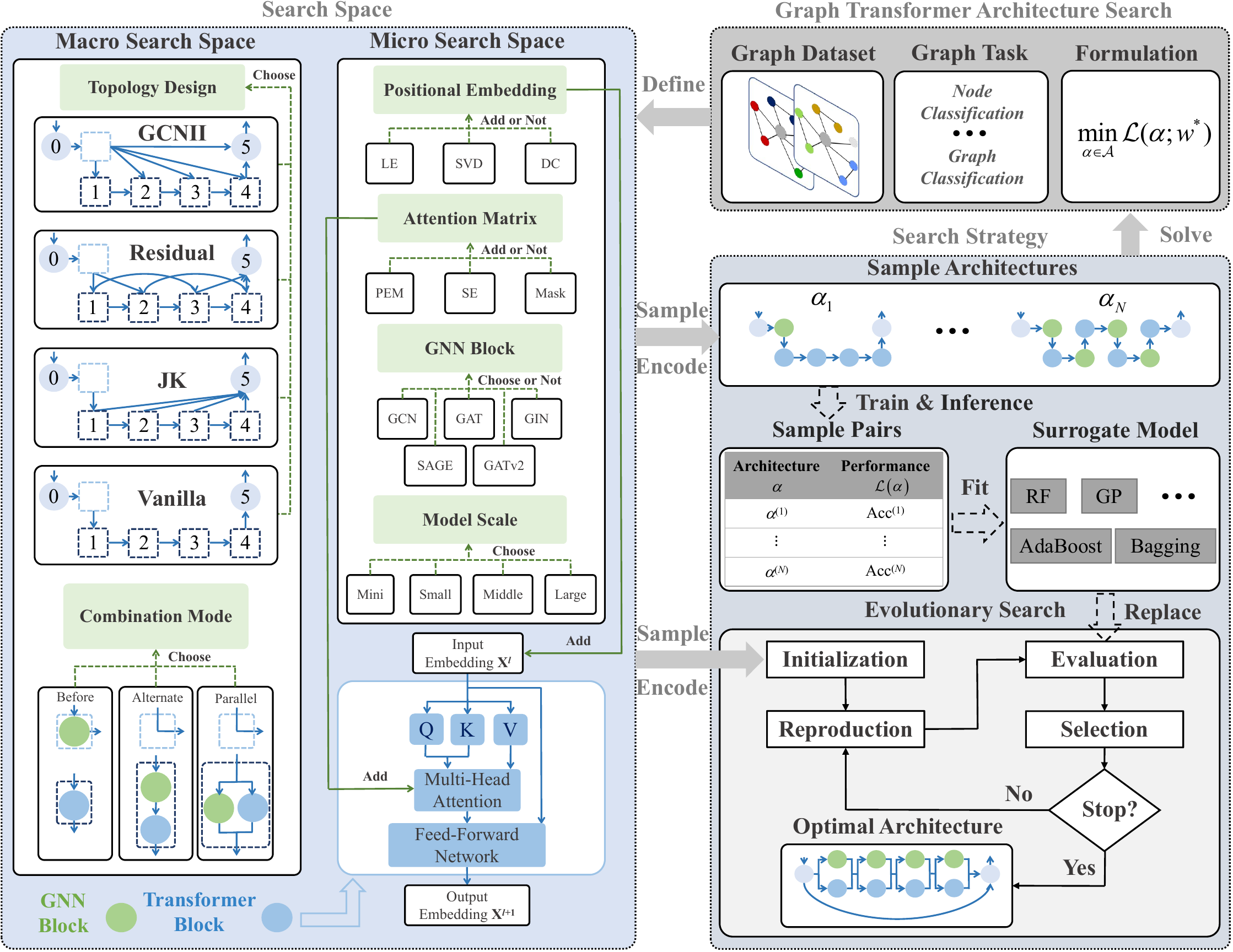}
\caption{\textcolor{black}{The overview of EGTAS consists of three components: problem definition, search space, and search strategy. According to the problem definition, the search space is carefully designed, encompassing both macro and micro levels. Then, a surrogate-assisted evolutionary search is presented to explore the search space to automate the construction of optimal graph Transformer architectures. Surrogate models are employed to predict the performance of architectures, thus reducing the computational cost of evaluations during the evolutionary search.}} \label{overview}
\end{figure*}

Configuring different topologies and graph-aware strategies can capture graph structure information at different levels to improve the Transformer's performance. This paper proposes an evolutionary graph Transformer architecture search (EGTAS) framework to tailor task-specific high-performance models automatically. As shown in Fig. \ref{overview}, EGTAS designs a comprehensive search space to unify topology design at the macro level and graph-aware strategies at the micro level, covering most of the advanced graph Transformer architectures. Then, we develop a surrogate-assisted evolutionary search for optimal architectural configurations by addressing obvious performance gaps induced by two levels of components in the search space. EGTAS presents a general architectural encoding covering topologies and graph-aware strategies. Since evaluating the performance of a candidate architecture is computationally expensive, a generic encoding-based surrogate model is constructed to predict architecture performance directly. Specifically, the model is trained on sample pairs consisting of the architectures and these performance metrics. The trained model is then used as the evaluator in the evolutionary search to speed up the optimization process. \textcolor{black}{To demonstrate the effectiveness of the EGTAS framework, extensive experiments are conducted on both benchmark and real-world graph datasets, encompassing node-level and graph-level tasks.} The performance gains demonstrate that EGTAS outperforms state-of-the-art GNAS and hand-crafted baselines. Ablation experiments show that the proposed surrogate models in evolutionary search can provide consistent performance predictions. We further demonstrate the necessity of joint exploration of graph Transformer components from macro and micro levels. In summary, the key contributions of this paper are as follows:

\begin{itemize}
  \item To the best of our knowledge, this is the first effort to study a unified search space to jointly explore model topologies and graph-aware strategies, which cover most state-of-the-art graph Transformer architectures.
  \item We introduce a simple but effective surrogate-assisted EGTAS framework to enable the automatic search for the best graph Transformer architecture. To speed up the evolutionary search process, a generic architectural encoding-based surrogate model is constructed to provide consistent performance predictions.
  \item \textcolor{black}{Extensive studies on benchmark and real-world graph datasets across representative tasks at both the node and graph levels demonstrate the scalability and practicality of EGTAS.} Ablation studies show that EGTAS can automatically construct architectures at different model scales, thereby alleviating the over-smoothing problem.
\end{itemize}

The rest of this paper is organized as follows. Section II presents related works on the graph neural architecture search and graph Transformer. \textcolor{black}{The working mechanism of the graph Transformer is introduced in Section III.} In Section IV, we describe the proposed EGTAS framework in detail. To validate the effectiveness and the efficiency of the EGTAS, a series of experiments are conducted in Section V along with the ablation study. Finally, we draw conclusions and introduce potential directions for future research in Section VI.

\section{Related Work}

A series of studies on GNAS have been proposed to automatically search model architectures for graph tasks in different computing scenarios. We first review related work on GNAS from the perspective of search space and search strategy. \textcolor{black}{Then, we introduce the key techniques in graph Transformer.}

\subsection{Graph Neural Architecture Search}
\subsubsection{Search Space} The search space can be divided into a macro search space and a micro search space. The macro search space determines the topology of GNNs. Wei \textit{et al.} \cite{10.1145/3485447.3512185} presented an F$^2$GNN framework to design topologies from a novel feature fusion perspective. Qin \textit{et al.} \cite{qin2022nasbenchgraph} developed a tailored benchmark, NAS-Bench-Graph, in which nine different topology designs are considered. On the other hand, the micro search space defines how information is exchanged across nodes in GNNs \cite{ijcai2020p195,10.3389/fdata.2022.1029307,10.1007/s10489-022-04096-w}. Common components include aggregation functions, aggregation weights, combination functions, hidden dimensions, activation functions, and so on. Nunes \textit{et al.} \cite{10.1145/3449639.3459318} analyzed a simple micro search space of GNAS using fitness landscape analysis, showing that the fitness landscape is easy to explore. In addition, a pooling search space is proposed to design GNNs for graph-level tasks adaptively \cite{9378060,NEURIPS2018_e77dbaf6,10.1145/3459637.3482285}.

\subsubsection{Search strategy} Common search strategies include reinforcement learning (RL), evolutionary algorithms (EA), and differentiable methods. RL-based methods \cite{zoph2017neural} train a controller to produce superior architectures, such as GraphNAS \cite{ijcai2020p195}, AGNN \cite{10.3389/fdata.2022.1029307}, and SNAG \cite{zhao2020simplifying}. Recently, Gao \textit{et al.} \cite{9782531} proposed a distributed GNAS framework that extends RL-based methods to distributed computing environments to improve search efficiency. In addition, Gao \textit{et al.} \cite{10040227} developed a generative adversarial learning framework aimed at training controllers for the automatic design of heterogeneous GNNs.

EA-based methods \cite{9508774,9954312,10132401} maintain an architecture population through some genetic operators (crossover, mutation, selection) to approximate the optimal architecture. The parent architecture in the population generates offspring architectures through crossover and mutation operators, and then those excellent architectures are selected to update the population. Shi \textit{et al.} \cite{SHI2022108752} first presented a co-evolutionary GNAS method to jointly search architecture and hyper-parameters. Chen \textit{et al.} \cite{9714826} proposed a parallel GNAS framework to improve search efficiency with the help of parallel computing technology. Wang \textit{et al.} \cite{10056291} developed a federated evolutionary strategy to solve the GNAS in the federated learning scenario, in which a super-network is used to speed up the evaluation of GNNs.

Differentiable methods \cite{liu2018darts} construct a super-network containing all candidate operations, where each operation is considered as a probability distribution over all candidate operations. These methods use a gradient-based algorithm to iteratively optimize both architecture and weights. For instance, Zhao \textit{et al.} \cite{9458743} first proposed a differentiable GNAS method, \textcolor{black}{SANE}, to search the macro-level and micro-level architecture of GNN. Qin \textit{et al.} \cite{NEURIPS2021_8c9f32e0} developed a differentiable graph architecture search with structure optimization, enabling the joint optimization of graph architecture and structure. Wei \textit{et al.} \cite{10.1145/3584945} presented a differentiable pooling architecture search to customize the GNN architecture for GC tasks. Wei \textit{et al.} \cite{10.1145/3485447.3512185} proposed \textcolor{black}{F$^2$GNN} to design the topology of GNNs, which shows state-of-the-art performance on graph learning tasks. \textcolor{black}{In \cite{guan2022large}, GAUSS is proposed to solve large-scale GNAS problems by introducing a lightweight supernet and joint architecture-graph sampling.}

\subsection{\textcolor{black}{Graph Transformer}}

\textcolor{black}{Designing Transformers for graph-structured data has received considerable attention \cite{min2022transformer}, aiming to alleviate the over-smoothing issue \cite{10068275}. Several studies have introduced graph information into positional encodings and attention matrices, allowing Transformers to understand graph structures. SAN \cite{kreuzer2021rethinking} designs a learnable positional encoding by utilizing the full Laplacian spectrum and incorporates real and virtual edge information into the attention mechanism. EGT \cite{hussain2021edge} presents a positional encoding based on the singular value decomposition (SVD) and proposes edge channels to focus on edge embeddings. Graphormer \cite{ying2021do} encodes various graph structure information into Transformer, exhibiting excellence in large-scale graph classification. GMT \cite{10.1145/3477495.3532031} explores graph masking mechanisms in the attention matrix, considering different types of interaction graphs. K-Subtree SAT \cite{chen2022structure} extracts a subgraph representation rooted at each node through existing GNNs before applying the attention mechanism. Rampášek \textit{et al.} \cite{rampavsek2022recipe} proposed a modular framework GPS that supports multiple types of graph encodings, showing competitive results in various benchmarks. In addition, Kong \textit{et al.} \cite{pmlr-v202-kong23a} presented a global Transformer to address node classification tasks on large-scale graphs.}

\section{\textcolor{black}{Preliminaries}}
A vanilla Transformer \cite{NIPS2017_3f5ee243} is stacked with multiple blocks, each of which consists of a multi-head attention (MHA) and feed-forward network (FFN). Given a node representation (input embedding) $\mathbf{X}$$^{l}\in\mathbb{R}^{n\times d}$, through block $l$, the new node representation $\mathbf{X}$$^{l+1}\in\mathbb{R}^{n\times d}$ can be expressed as:

\begin{align}
\mathbf{A}^{l}_{h} & =\operatorname{SoftMax}\left(\frac{ \mathbf{X}^{l} \mathbf{Q}^{l}_{h}(\mathbf{X}^{l} \mathbf{K}^{l}_{h})^{\top}}{\sqrt{d}}\right), \label{MHA_A}\\
\mathbf{O}^{l} & = \parallel_{h=1}^H\left(\mathbf{A}^{l}_{h}\mathbf{X}^{l}\mathbf{V}^{l}_{h}\right)\mathbf{W_O}^{l} + \mathbf{X}^{l},\label{MHA_B}\\
\mathbf{X}^{l+1} & =\sigma\left(\mathbf{O}^{l} \mathbf{W}_1^{l}+\mathbf{b}_1^{l}\right) \mathbf{W}_2^{l}+\mathbf{b}_2^{l}, \label{FFN}
\end{align}
where Eq. (\ref{MHA_A}) and Eq. (\ref{MHA_B}) are the MHA component, while Eq. (\ref{FFN}) is the FFN component.  $\operatorname{SoftMax}$ represents a softmax function. $\mathbf{A}^{l}_{h}\in \mathbb{R}^{n\times n}$ and $\mathbf{O}^{l}\in\mathbb{R}^{n\times d}$ are the message-passing matrix and output of $H$ multi-head attention mechanisms, respectively. $\mathbf{Q}^{l}_{h},\mathbf{K}^{l}_{h},\mathbf{V}^{l}_{h}\in \mathbb{R}^{d\times d_h},\mathbf{W_O}^{l}\in \mathbb{R}^{(H\times d_h)\times d},\mathbf{W}_1^{l}\in \mathbb{R}^{d\times d_1},\mathbf{b}_1^{l}\in \mathbb{R}^{d_1},\mathbf{W}_2^{l}\in \mathbb{R}^{d_1\times d},\mathbf{b}_2^{l}\in \mathbb{R}^{d}$ are the learnable parameters. For graph learning tasks, all node features $\mathbf{X}^{0}=\left[\mathbf{x}_1,...,\mathbf{x}_n\right]$ are employed as initial input. After going through $L$ blocks, we can obtain the final high-level node representation $\mathbf{X}^{L}$. 

\textcolor{black}{Next, we introduce the working mechanism of graph Transformers from the perspective of graph-aware strategies, which are closely aligned with the proposed EGTAS.} Typical graph-aware strategies for incorporating graph information into Transformers can be divided into three groups: 1) a combination of GNNs and Transformers (CGT), 2) positional embedding with graph information (PE), and 3) attention matrix with graph information (AM). It's worth noting that these strategies can be flexibly combined \cite{ying2021do,rampavsek2022recipe}.

\subsection{CGT} The basic idea of CGT is to combine GNNs and Transformers to process graph-structured data. According to the relative positions of GNNs and Transformer, existing CGT-based methods include three types: before, alternate, and parallel. For the before type, a Transformer is added on top of a GNN layer \cite{jain2021representing, NEURIPS2020_94aef384,chen2022structure}. GNN blocks are employed to learn high-level representations of a node neighborhood, while Transformer blocks learn all pairwise node interactions in a position-agnostic manner. For the alternate type, GNN blocks and Transformer blocks are stacked alternately. For instance, Lin \textit{et al.} \cite{Lin_2021_ICCV} stacked a Transformer module consisting of a graph residual block and a multi-head self-attention layer. For the parallel type, GNN blocks and Transformer blocks are designed to work in parallel. Zhang \textit{et al.} \cite{zhang2020graph} proposed a Graph-BERT architecture in which a graph residual term is added to each basic attention layer.

\subsection{PE} The second graph-aware strategy compresses the graph structure into positional embedding vectors and adds graph information to the input block before the Transformer model:
\begin{equation}
    \mathbf{X}_{PE}^{l}=\mathbf{X}^{l}+f_{PE}(P_G),
\end{equation}
where $P_G\in \mathbb{R}^{n\times d_p}$ is the graph embedding vectors, and $f_{PE}: \mathbb{R}^{n\times d_p} \rightarrow{\mathbb{R}^{n\times d}}$ represents a simple feed-forward neural network to align the dimensions of vectors. Existing works adopt various methods as graph embedding vectors \cite{kreuzer2021rethinking}, such as Laplacian eigenvectors (LE) \cite{dwivedi2020generalization} and SVD-based positional encodings \cite{hussain2021edge}. Different from the compressed adjacency matrix in the above methods, Ying \textit{et al.} \cite{ying2021do} proposed a heuristic method (called DC in this paper) to extract structural information from the adjacency matrix, where the out-degree and in-degree of each node are used to design embedding vectors. By utilizing degree centrality, Transformers can capture key information of nodes.

\subsection{AM} The third graph-aware strategy generates an additional graph-based attention mechanism $f_{AM}(P)$ to enhance the original attention matrix:
\begin{equation}
    \mathbf{A}^{l}_{h,AM} =\operatorname{SoftMax}\left(\frac{ \mathbf{X}^{l} \mathbf{Q}^{l}_{h}(\mathbf{X}^{l} \mathbf{K}^{l}_{h})^{\top}}{\sqrt{d}} + f_{AM}(P)\right),
\end{equation}
where $P$ is the graph information. Ying \textit{et al.} \cite{ying2021do} presented a novel spatial encoding-based attention mechanism (SE) that measures the spatial relation between nodes. To capture fine-grained structural information, Zhao \textit{et al.} \cite{zhao2021gophormer} proposed a proximity-enhanced multi-head attention mechanism (PEM). Min \textit{et al.} \cite{10.1145/3477495.3532031} developed a masking mechanism-based graph encoding strategy, where the graph-masking mechanism enforces heads to attend to different subspaces, thereby improving the representative capability of models.

\begin{table}[t]
\color{black}
\centering
\caption{Comparison of related works and proposed EGTAS from an architectural perspective.}\label{related_diff}
\resizebox{0.48\textwidth}{!}{
\begin{tabular}{cccccccc}
\toprule
Category & Methods & GNN Block & Topology & CGT & PE & AM & Model Scale \\
\midrule
\multirow{6}{*}{\makecell{Graph \\ Transformer}} & SAN \cite{kreuzer2021rethinking} & & & &\checkmark & \checkmark& \\
& EGT \cite{hussain2021edge} & & & & \checkmark & \checkmark & \\
& Graphormer \cite{ying2021do} &  & & \checkmark& \checkmark & \checkmark & \\
& K-Subtree SAT \cite{chen2022structure} & & \checkmark & \checkmark & \checkmark & & \\
& GMT \cite{10.1145/3477495.3532031} &  & & & & \checkmark & \\
& GPS \cite{rampavsek2022recipe} & & & \checkmark & \checkmark & \checkmark & \\
\midrule
\multirow{6}{*}{\makecell{GNAS}} & SNAG \cite{zhao2020simplifying} & \checkmark & \checkmark & & & & \\
& SANE \cite{9458743} & \checkmark & \checkmark & & & &  \\
& F$^2$GNN \cite{10.1145/3485447.3512185} & \checkmark & \checkmark & & & & \\
& GAUSS \cite{guan2022large} & \checkmark & & & & & \\
& AutoGT \cite{zhang2023autogt} & & & & \checkmark & \checkmark & \checkmark \\
& \textbf{Our proposed EGTAS} & \checkmark& \checkmark& \checkmark& \checkmark & \checkmark & \checkmark \\
\bottomrule
\end{tabular}}
\end{table}

\subsection{\textcolor{black}{Discussion}}

\textcolor{black}{Table \ref{related_diff} summarizes state-of-the-art GNAS methods and graph Transformers from an architectural perspective. Compared to existing works, EGTAS introduces a comprehensive search space to jointly explore both macro and micro designs of graph transformers. The macro designs focus on topology and CGT, while the micro ones encompass GNN blocks, PE, AM, and model scale.}

\section{EGTAS: Evolutionary Graph Transformer Architecture Search}

The overall framework of EGTAS is presented in Fig. \ref{overview}, which consists of three parts: problem definition, search space, and search strategy. Next, we introduce their technical details.

\subsection{Problem Definition}
\begin{definition}
\label{def_GNAS}
(\textit{\textbf{GNAS}}) 
In general, given a graph dataset $\mathcal{D} = \{\mathcal{D}_{tra}, \mathcal{D}_{val}, \mathcal{D}_{test}\}$ drawn from a data distribution $P_\mathcal{D}$, a simple GNAS problem is described as:
\end{definition}
\begin{equation}\label{GNAS_eq}
\begin{split}
\underset{\bm{\alpha} \in \mathcal{A}}{\operatorname{minimize}} & \ \mathcal{L}(\bm{\alpha}; \bm{w}^*)\\
s.t. &\ \bm{w}^*(\bm{\alpha})\in \underset{\bm{w} \in \mathcal{W}}{\operatorname{argmin}} \ \mathcal{F}(\bm{w};\bm{\alpha})\\
\end{split},
\end{equation}
where $\bm{\alpha}$ and $\bm{w}$ define the architecture and its weights, respectively. $\mathcal{F}$ is the loss function on the training set $\mathcal{D}_{tra}$ for a candidate architecture $\bm{\alpha}$. $\mathcal{L}$ is the upper-level objective on the validation set $\mathcal{D}_{val}$, such as \textcolor{black}{the mean absolute error (\#MAE), accuracy (\#Acc), and area under the curve (\#AUC)}.

In typical lower-level optimization, learning the weights of candidate architectures requires resource-intensive stochastic gradient descents spanning multiple epochs. Therefore, the GNAS problem is computationally expensive. \textcolor{black}{Next, we illustrate the specific problem formulations of GNAS through two representative tasks: NC and GC.}

Given a training dataset $\mathcal{D}_{tra}=\{\left(\mathbf{X},\mathbf{y}\right)\}$ with node-level features $\mathbf{X}=\{\mathbf{X}_1,...,\mathbf{X}_{n_c}\}$ and their labels $\mathbf{y}=\{\mathbf{y}_1,...,\mathbf{y}_{n_c}\}$, NC aims to predict the labels of nodes in a graph $\mathbf{G}$ by optimizing the negative log-likelihood loss:
\begin{equation}\label{NC_LL}
\mathcal{F}_{NC}(\bm{w};\bm{\alpha}) = \sum_{i=1}^{n_c}{-log(\operatorname{SoftMax}(f_{\bm{w};\bm{\alpha}}(\mathbf{X}_{i}),\mathbf{y}_{i}))},
\end{equation}
\textcolor{black}{where $f_{\bm{w};\bm{\alpha}}$ represents the output of the architecture $\bm{\alpha}$ with weights $\bm{w}$.}
\begin{definition}
\label{def_NC}
(\textit{\textbf{GNAS for NC}}) 
Given a validation dataset $\mathcal{D}_{val}=\{(\mathbf{X}^{'},\mathbf{y}^{'})\}$ with node-level features $\mathbf{X}^{'}=\{\mathbf{X}^{'}_1,...,\mathbf{X}^{'}_{n_{cv}}\}$ and their labels $\mathbf{y}^{'}=\{\mathbf{y}^{'}_1,...,\mathbf{y}^{'}_{n_{cv}}\}$, the GNAS problem for NC can be formalized as:
\begin{equation}\label{GNAS_NC}
\begin{split}
\underset{\bm{\alpha} \in \mathcal{A}}{\operatorname{minimize}} & \ \mathcal{L}_{NC}(\bm{\alpha}; \bm{w}^*)= \sum_{i=1}^{n_{cv}}{l_{NC}(f_{\bm{\alpha}; \bm{w}^*}(\mathbf{X}^{'}_i),\mathbf{y}^{'}_i)}\\
s.t. &\ \bm{w}^*(\bm{\alpha})\in \underset{\bm{w} \in \mathcal{W}}{\operatorname{argmin}} \ \mathcal{F}_{NC}(\bm{w};\bm{\alpha})\\
\end{split},
\end{equation}
where $f_{\bm{\alpha};\bm{w}^*}$ represents the output of the architecture $\bm{\alpha}$ with optimal weights $\bm{w}^*$. \#Acc is employed as the upper-level objective $l_{NC}$ in this paper.
\end{definition}

Given a training dataset $\mathcal{D}_{tra}=\{\left(\mathbf{X}_{\mathbf{G}},\mathbf{y}_{\mathbf{G}}\right)\}$ with graph-level features $\mathbf{X}_{\mathbf{G}}=\{\mathbf{X}_{\mathbf{G}^1},...,\mathbf{X}_{\mathbf{G}^n}\}$ and their labels $\mathbf{y}_{\mathbf{G}}=\{\mathbf{y}_{\mathbf{G}^1},...,\mathbf{y}_{\mathbf{G}^n}\}$, GC aims to predict labels of all graphs $\{\mathbf{G}^1,...,\mathbf{G}^n\}$ by minimizing the squared-error loss:
\begin{equation}\label{GC_LL}
\mathcal{F}_{GC}(\bm{w};\bm{\alpha}) = \sum_{i=1}^n {||f_{\bm{w};\bm{\alpha}}(\mathbf{X}_{\mathbf{G}^i})-\mathbf{y}_{\mathbf{G}^i}||_2^2}.
\end{equation}

\begin{definition}
\label{def_GC}
(\textit{\textbf{GNAS for GC}}) 
Given a validation dataset $\mathcal{D}_{val}=\{(\mathbf{X}^{'}_{\mathbf{G}},\mathbf{y}^{'}_{\mathbf{G}})\}$ with graph-level features $\mathbf{X}^{'}_{\mathbf{G}}=\{\mathbf{X}^{'}_{\mathbf{G}^1},...,\mathbf{X}^{'}_{\mathbf{G}^{n_v}}\}$ and their labels $\mathbf{y}^{'}_{\mathbf{G}}=\{\mathbf{y}^{'}_{\mathbf{G}^1},...,\mathbf{y}^{'}_{\mathbf{G}^{n_v}}\}$, the GNAS problem for GC is:
\begin{equation}\label{GNAS_GC}
\begin{split}
\underset{\bm{\alpha} \in \mathcal{A}}{\operatorname{minimize}} & \ \mathcal{L}_{GC}(\bm{\alpha}; \bm{w}^*)= \sum_{i=1}^{n_{v}}{l_{GC}(f_{\bm{\alpha}; \bm{w}^*}(\mathbf{X}^{'}_{\mathbf{G}^i}),\mathbf{y}^{'}_{\mathbf{G}^i})}\\
s.t. &\ \bm{w}^*(\bm{\alpha})\in \underset{\bm{w} \in \mathcal{W}}{\operatorname{argmin}} \ \mathcal{F}_{GC}(\bm{w};\bm{\alpha})\\
\end{split},
\end{equation}
where the upper-level objective $l_{GC}$ is \#Acc or \#AUC.
\end{definition}

\subsection{Comprehensive Graph Transformer Search Space}
\begin{figure}[t]
\centering
\includegraphics[width=0.4\textwidth]{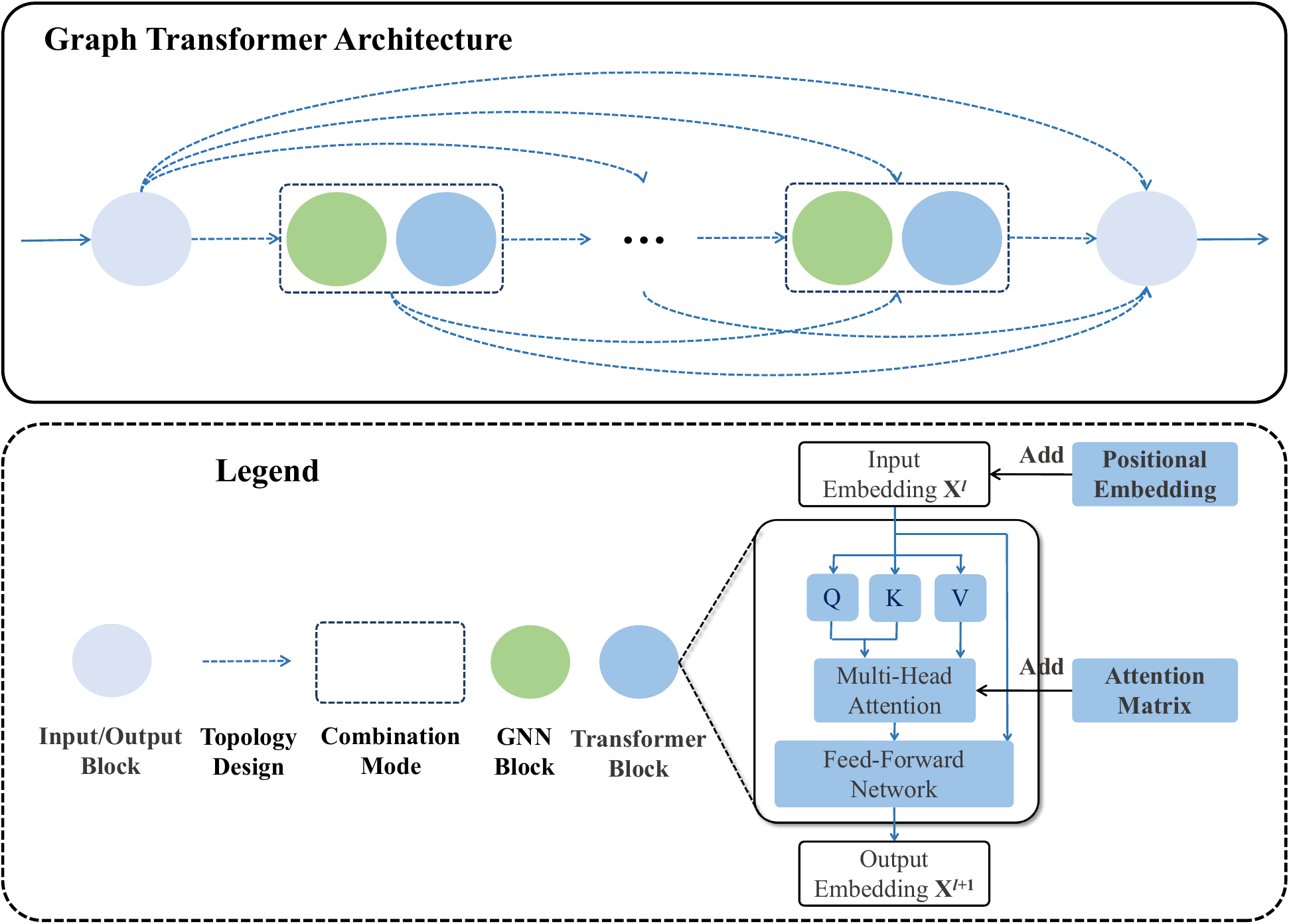}
\caption{\textcolor{black}{A general architecture of the graph Transformer.}} \label{trans-arch}
\end{figure}

\textcolor{black}{Fig. \ref{trans-arch} showcases the general architecture of the graph Transformer, which is constructed by connecting multiple graph Transformer blocks in various topological manners. Existing methods mainly focus on designing graph-aware strategies to integrate graph information into Transformer blocks. Each graph Transformer block flexibly combines a GNN and a Transformer block. For the Transformer block, graph information is fused into the positional embedding and attention matrix. In summary, topology and graph-aware strategies jointly determine a graph Transformer architecture.}

A suitable architecture search space should be diverse and compact, which allows the space to cover powerful architectures and guarantees efficient searches. Keeping this principle in mind, we propose the unified search space $\mathcal{A}$ at both macro and micro levels as shown in Fig. \ref{overview} (left) and Table \ref{MSS}. In the unified search space, we provide a set of candidate operations for topologies and graph-aware strategies.

\subsubsection{Macro Search Space}
The macro search space reveals topological patterns of graph Transformers that need to be explored. Table \ref{MSS} lists the operations on the macro search space, including the topology design and the combination mode of \textcolor{black}{GNN and Transformer blocks}.
\begin{table}[t]
\centering
\caption{Operations used in our search space $\mathcal{A}$.}\label{MSS}
\resizebox{0.48\textwidth}{!}{
\begin{tabular}{lll}
\toprule
Levels & Operations & Values \\
\midrule
\multirow{2}{*}{Macro} & Topology Design & Vanilla, JK, Residual, GCNII  \\
\cmidrule{2-3}
& Combination Mode & Before, Alternate, Parallel \\
\midrule
\multirow{7}{*}{Micro} & GNN Block & GCN, SAGE, GAT, GATv2, GIN, None \\
\cmidrule{2-3}
& Positional Embedding & \makecell[l]{None, [LE], [SVD], [DC], [LE,SVD], \\ {[LE,DC]}, [SVD,DC], [LE,SVD,DC]} \\
\cmidrule{2-3}
& Attention Matrix & \makecell[l]{None, [PEM], [SE], [Mask], [PEM,SE], \\ {[PEM,Mask]}, [SE,Mask], [PEM,SE,Mask]} \\
\cmidrule{2-3}
& Model Scale & Mini, Small, Middle, Large \\
\bottomrule
\end{tabular}}
\end{table}

\paragraph{\textcolor{black}{Topology Design}} Graph Transformers with high model capacity can be built by designing topology. Common topology designs are mainly presented in two paradigms: stack operation and skip connection. The former stacks blocks to capture higher-level graph features. The latter integrates features at different levels via skip connections to provide diverse local feature extraction. We provide four popular topology designs in the macro search space as shown in Fig. \ref{overview} (left), i.e., Vanilla, JK \cite{Li_2019_ICCV}, Residual \cite{pmlr-v80-xu18c}, and GCNII \cite{pmlr-v119-chen20v}. Without loss of generality, other topology designs can also be easily integrated with the proposed framework. Given the input $\mathbf{X}$ of a graph Transformer model $f$ with $L$ blocks, the output of $f$ via the above topology design can be formalized as follows:
\begin{align}
\color{black}
\text{Vanilla}&:\quad \mathbf{X}^{l+1} =f_l(\mathbf{X}^{l}), \label{Vanilla}\\
\color{black}
\text{JK}&:\quad \mathbf{X}^{L} =f(\mathbf{X}^{0},...,\mathbf{X}^{L-1}),\label{JK}\\
\color{black}
\text{Residual}&:\quad \mathbf{X}^{l+1} =f_l(\mathbf{X}^{l} + \mathbf{X}^{l-1}),\label{Residual}\\
\color{black}
\text{GNNII}&:\quad \mathbf{X}^{l+1} =\alpha \mathbf{X}^{0} + (1-\alpha) f_l(\mathbf{X}^{l}).\label{GCNII}
\end{align}

Eq. (\ref{Vanilla}) represents the vanilla paradigm that is constructed by stacking multiple operations. The JK and Residual paradigms adopt multiple skip connection modes to utilize different levels of features, as shown in Eq. (\ref{JK}) and (\ref{Residual}). Eq. \ref{GCNII} indicates that GNNII adds initial features to each block to alleviate the over-smoothing problem. 

\paragraph{\textcolor{black}{Combination Mode}} In the macro search space, the graph Transformer block is composed of \textcolor{black}{GNN and Transformer blocks}, as shown in Fig. \ref{overview} (left). The combination mode is designed in three major manners: Before \cite{jain2021representing}, Alternate \cite{Lin_2021_ICCV}, and Parallel \cite{zhang2020graph}. 

\textcolor{black}{Before: }The Before paradigm is widely adopted in the existing literature \cite{jain2021representing}, where the GNN block is added in front of the vanilla Transformer. GNNs can learn the local representation of nodes, while the Transformer is employed to enhance the global reasoning ability of the model.

\textcolor{black}{Alternate: }For the Alternate paradigm, each graph Transformer block is stacked by a GNN block and a Transformer block, where each block considers both local and global information. 

\textcolor{black}{Parallel: } In the Parallel paradigm, a Transformer block and a GNN block extract high-level features in parallel to improve model capacity.

\subsubsection{Micro Search Space}
Table \ref{MSS} shows the micro search space, including the GNN block, the positional embedding with graph information, the attention matrix with graph information, and the model scale. In the micro search space, we mainly explore the design of graph Transformer blocks, which determines how nodes exchange information with other nodes in each block. The graph Transformer block consists of a GNN block and a Transformer block. For the Transformer block, two graph-aware strategies are considered, i.e., designing position embedding and attention matrix. It is worth noting that each operation in the graph-aware strategies is further decided whether to use it. \textcolor{black}{Next, we introduce the GNN block, the positional embedding with graph information, and the attention matrix with graph information in detail.}

\paragraph{\textcolor{black}{GNN Block}} For the GNN block, we consider the six popular choices \cite{JMLR:v23:20-852,qin2022nasbenchgraph}: GCN \cite{kipf2017semisupervised}, SAGE \cite{hamilton2017inductive}, GAT \cite{veličković2018graph}, GATv2 \cite{brody2022how}, GIN \cite{xu2018how}, and None. None means that the GNN block is not adopted in the graph Transformer blocks. \textcolor{black}{GCN represents a localized first-order approximation of a spectral filter on graphs. It efficiently learns feature representations from graph data by defining convolution operations specifically for graphs. SAGE learns embeddings for unseen nodes on large graphs by sampling and aggregating features from a node's local neighborhood. GAT incorporates an attention mechanism into GNNs. It assigns different attention weights to neighboring nodes, dynamically focusing on the most relevant graph information during propagation. Compared to GAT, GATv2 enhances the expressiveness of the attention mechanism, allowing every node to attend to any other node. GIN mixes the features obtained after each aggregation operation with the original node features, aiming to ensure the model's sensitivity to graph isomorphism. These GNN blocks propagate feature information in different aggregation manners between neighboring nodes to learn deeper feature representations. All the above GNN blocks are implemented by PyTorch Geometric with the default settings \cite{fey2019fast}.
}

\paragraph{\textcolor{black}{Positional Embedding}}  For the positional embedding, three candidate operations are considered including LE \cite{dwivedi2020generalization}, SVD \cite{hussain2021edge}, and DC \cite{ying2021do}. Notice that each operation has to be considered whether to add to the positional embedding. 

\textcolor{black}{LE: } LE \cite{dwivedi2020generalization} adopts eigenvectors of the $k$ smallest non-trivial eigenvalues as graph information, which are concatenated to the input node attribute matrix. These eigenvectors are obtained by factoring the graph Laplace matrix:
\begin{equation} \label{LE}
    \mathbf{U}^T \Lambda \mathbf{U}=\mathbf{I}-\mathbf{D}^{-1 / 2} \mathbf{A} \mathbf{D}^{-1 / 2},
\end{equation}
where $\mathbf{A}$ and $\mathbf{D}$ are the adjacent matrix and degree matrix, respectively. $\mathbf{U}$ and $\Lambda$ are accordingly eigenvectors and eigenvalues.

\textcolor{black}{SVD: } The SVD-based operation \cite{hussain2021edge} uses left and right singular vectors of the largest $k$ singular values as graph information. These vectors can be calculated by conducting singular value decomposition to the graph adjacent matrix $\mathbf{A}$:
\begin{equation} \label{SVD}
    \mathbf{A}{\approx} \mathbf{U} \boldsymbol{\Sigma} \mathbf{V}^T=(\mathbf{U} \sqrt{\boldsymbol{\Sigma}}) \cdot(\mathbf{V} \sqrt{\boldsymbol{\Sigma}})^T=\tilde{\mathbf{U}} \tilde{\mathbf{V}}^T,
\end{equation}
where $\mathbf{U},\mathbf{V} \in \mathbb{R}^{n\times k}$ contains the left and right singular vectors of the largest $k$ singular values. Similar to LE, we concatenate these vectors to the input node attribute matrix. 

\textcolor{black}{DC: } The DC-based operation \cite{ying2021do} constructs two real-valued embedding vectors according to the in-degree and out-degree of nodes as graph information. Therefore, the input embedding $\tilde{\mathbf{X}}_i^{l}$ of node $i$ in the $l$ block can be formalized as:
\begin{equation} \label{PE}
\tilde{\mathbf{X}}_i^{l}=\mathbf{X}_i^{l}+\mathbf{z}^{-}_{v_i}+\mathbf{z}^{+}_{v_i},
\end{equation}
where $\mathbf{z}^{-}_{v_i}$ and $\mathbf{z}^{+}_{v_i}$ are the embedding vectors generated by the in-degree and out-degree of the node $v_i$, respectively.

\paragraph{\textcolor{black}{Attention Matrix}}  For the attention matrix with graph information, we adopt three popular operations: PEM \cite{zhao2021gophormer}, SE \cite{ying2021do}, and Mask \cite{10.1145/3477495.3532031}. Similar to the positional embedding, each operation is considered whether to add to the attention matrix. 

\textcolor{black}{PEM: } For a node pair $(v_i,v_j)$, PEM-based operation \cite{ying2021do} adds multiple structural information to the attention matrix:
\begin{equation} \label{PEM}
\mathbf{A}^{l}_{ij} =\operatorname{SoftMax}\left(\frac{ \mathbf{X}_{i}^{l} \mathbf{Q}^{l}(\mathbf{X}_{j}^{l} \mathbf{K}^{l})^{\top}}{\sqrt{d}} + \psi^{i j} \mathbf{b}^{\top}\right),
\end{equation}
where $\mathbf{b}$ is the learnable parameters. The structural encoding $\psi^{i j}$ can be generated by $n$ structural functions $\Psi_n\left(\cdot \right)$: $\psi_{i j}=\operatorname{Concat}\left(\Psi_n\left(v_i, v_j\right) \mid n \in 0,1, \ldots, n-1\right)$. 

\textcolor{black}{SE: } SE-based operation \cite{ying2021do} adds spatial encoding to the attention matrix:
\begin{equation} \label{SE}
\mathbf{A}^{l} =\operatorname{SoftMax}\left(\frac{ \mathbf{X}^{l} \mathbf{Q}^{l}(\mathbf{X}^{l} \mathbf{K}^{l})^{\top}}{\sqrt{d}} + \mathbf{B}_{se}\right),
\end{equation}
where $\mathbf{B}_{se}$ is a bias matrix generated by the shortest path distance $\psi$ between nodes.

\textcolor{black}{Mask: } For a node pair $(v_i,v_j)$, Mask-based operation \cite{10.1145/3477495.3532031} uses graph masking mechanism to improve the attention matrix:
\begin{align} \label{Mask}
    \mathbf{A}^{l}_{ij} &=\operatorname{SoftMax}\left(\frac{ \mathbf{X}_{i}^{l} \mathbf{Q}^{l}(\mathbf{X}_{j}^{l} \mathbf{K}^{l})^{\top}}{\sqrt{d}} + z_m^{i,j}\right),\\
    z_m^{i,j} &= \begin{cases}
0 & \text{ if } \psi(v_i,v_j) \leq m \\
-\infty & \text{ else } 
\end{cases},
\end{align}
where $z_m^{i,j}$ reflects the relationship between the shortest path distance $\psi(v_i,v_j)$ and a threshold parameter $m$. Attention is masked when the given threshold is not met.

\begin{table}[t]
\centering
\caption{The model scale patterns describing key hyper-parameters of graph Transformer sizes.}\label{model_scale}
\begin{tabular}{ccccc}
\toprule
\multirow{2}{*}{Hyper-parameters} & \multicolumn{4}{c}{Model Scale} \\
 & Mini & Small & Middle & Large \\
\midrule
 Layer & 3 & 6 & 12 & 12  \\
 Hidden Dimension & 64 & 80 & 80 & 512 \\
 Attention Head & 4 & 8 & 8 & 32 \\
 Attention Head Dimension & 5 & 10 & 10 & 16 \\
 FFN Hidden Dimension & 64 & 80 & 80 & 512 \\
\bottomrule
\end{tabular}
\end{table}

\paragraph{\textcolor{black}{Model Scale}} As shown in Table \ref{model_scale}, we also define four model scale patterns describing key hyper-parameters of graph Transformer sizes based on empirical values \cite{min2022transformer}. These hyper-parameters include layer, hidden dimension, attention head, attention head dimension, and FFN hidden dimension. 

\subsection{Search Strategy}
Given the comprehensive search space in Section IV.B, a large number of operation combinations need to be explored by a search strategy. As indicated by \textbf{Definition} \ref{def_GNAS}, direct acquisition of gradient information pertaining to upper-level objectives (such as \#Acc) concerning the architecture is infeasible. Therefore, GNAS is usually regarded as a black-box optimization problem \cite{ijcai2021p637}. Due to excellent parallelism and gradient-free properties, evolutionary search is widely employed to solve black-box optimization problems \cite{9508774}. Next, we present the proposed surrogate-assisted evolutionary search from encoding strategy, surrogate model construction, and evolutionary search.

\begin{figure}[t]
\centering
\includegraphics[width=0.25\textwidth]{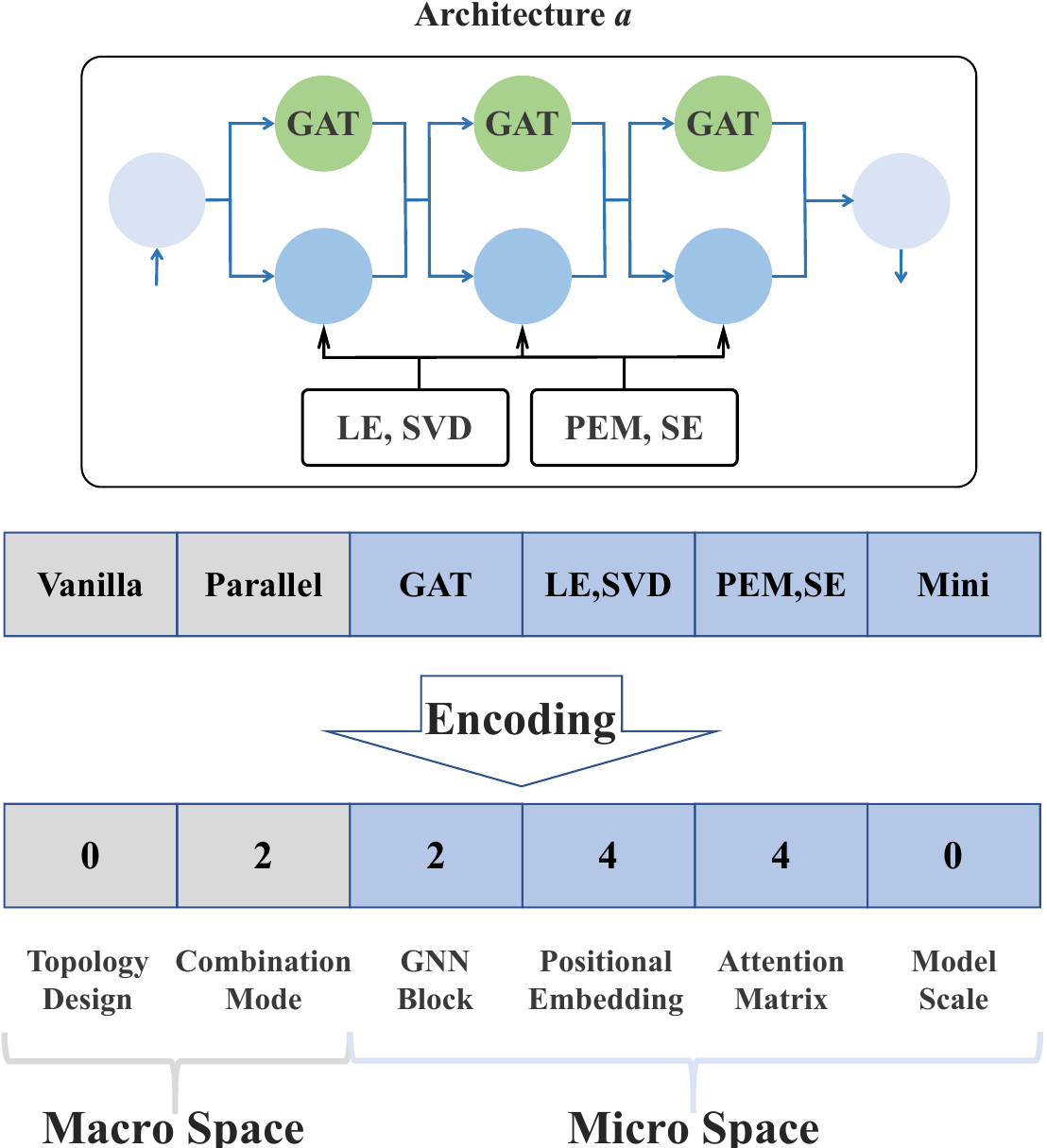}
\caption{\textcolor{black}{An illustrative example of the encoding strategy.}} \label{Encoding}
\end{figure}

\subsubsection{Encoding Strategy}
Evolutionary search needs to decide their encoding strategy at the beginning stage. This section introduces how to encode an architecture into an individual for the evolutionary search. A graph Transformer architecture $\alpha$ can be represented by an integer vector (chromosome) consisting of topology design, combination mode, positional embedding, attention matrix, GNN block, and model scale. Fig. \ref{Encoding} gives an illustrative example of a graph Transformer architecture. The chromosome value represents the index of the selected operation in Table \ref{MSS}. For example, the candidate operation Mini is encoded as 0 for the model scale. Notice that the encoding strategy is generic. New search components can be easily integrated into our method without loss of generality.

\begin{figure}[t]
\centering
\includegraphics[width=0.35\textwidth]{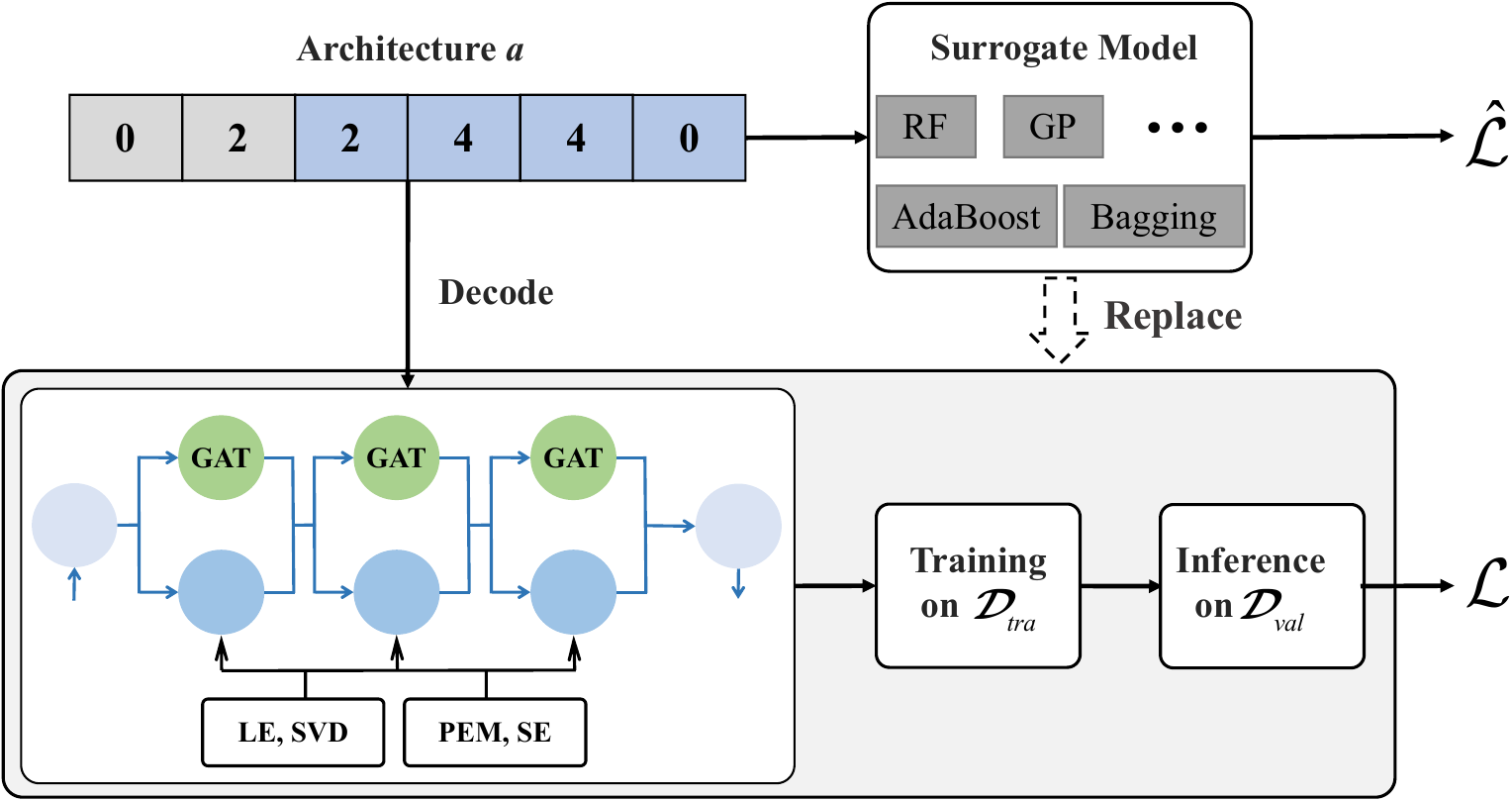}
\caption{\textcolor{black}{Evaluation process of an architecture $\alpha$.}} \label{eval_arches}
\end{figure}

\subsubsection{Surrogate Model Construction}
\textcolor{black}{Fig. \ref{eval_arches} illustrates the evaluation process of an architecture $\alpha$. According to Eq. \ref{GNAS_eq}, the actual evaluation process of each architecture needs to complete an expensive lower-level optimization involving decoding, model training, and inference. To reduce the computational cost, we employ the surrogate model to replace the actual evaluation process for rapid prediction of architecture performance.} The surrogate model fits the functional relationship between architecture and performance \cite{9723446}, which meets the following basic requirements: 1) low computational cost; 2) high-order correlation across predicted and real evaluations.

\begin{algorithm}[t]
 \caption{Surrogate Model Construction (SMC)} 
 \label{alg1}
 \begin{algorithmic}[1]
  \REQUIRE $\mathcal{D}_{tra}$: Training set; $\mathcal{D}_{val}$: Validation set; $N_s$: Number of samples.
  \ENSURE $M$: Surrogate model.
  \STATE  $\{\bm{\alpha}^{(i)}, i=1, ..., N_s\}\leftarrow$ Randomly sample $N_s$ architectures from $\mathcal{A}$ and encode them as vector sets;
  \STATE $\{\bm{w}^{(i)}, i=1,...,N_s\}\leftarrow$ Train $N_s$ architectures on the training set over multiple epochs;
  \STATE $\{\mathcal{L}^{(i)}(\bm{\alpha}^{(i)};\bm{w}^{(i)}),i=1,...,N_s\}\leftarrow$ Calculate the performance metrics on the validation set;
  \STATE $\{M^{(1)},...,M^{(m)}\} \leftarrow$ Fit $m$ surrogate models based on the training data $\{(\bm{\alpha}^{(i)} ,\mathcal{L}^{{(i)}}),i=1,...,N_s$;
  \STATE $M \leftarrow$ Select the best surrogate model from $\{M^{(1)},...,M^{(m)}\}$ via cross-validation.
 \end{algorithmic}
\end{algorithm}

The construction process of the whole surrogate model is shown in Algorithm \ref{alg1}. First, we randomly sample an architecture set from the search space \textcolor{black}{$\mathcal{A}$} and encode them as vector sets according to the encoding strategy. Gradient-based methods are used to train the architecture over multiple epochs. Then we compute performance metrics on the validation set. Finally, numerous surrogate models \textcolor{black}{$\{M^{(1)},...,M^{(m)}\} $} are trained based on the training data. \textcolor{black}{As no single surrogate model consistently outperforms others across all datasets \cite{585893}, we construct multiple models. Then the best model $M$ is selected based on the evaluation metric of MSE through cross-validation.} Six popular surrogate models are adopted including Decision Tree (DT), Random Forest (RF), AdaBoost, Bagging, Gaussian Process (GP), and ExtraTree \cite{Liu_2021_ICCV}. All models are implemented by Scikit-learn with the default settings \cite{JMLR:v12:pedregosa11a}.

\begin{algorithm}[t]
 \caption{Framework of EGTAS} 
 \label{alg2}
 \begin{algorithmic}[1]
  \REQUIRE $\mathcal{D}_{tra}$: Training set; $\mathcal{D}_{val}$: Validation set; $N_s$: Number of samples; $T$: Number of iterations; $N_p$: Population size; \textcolor{black}{$p_c$: Crossover probability;  $p_m$: Mutation probability.}
  \ENSURE $\bm{\alpha}^*,\bm{w}^*$: Best architecture and its weights.
  \STATE $M\leftarrow SMC(N_s, \mathcal{D}_{tra}, \mathcal{D}_{val})$;  // \textit{Algorithm \ref{alg1}}
  \STATE $i,t \leftarrow 0$ // \textit{Initialize counters}
  \STATE $P,\mathcal{\hat{L}} \leftarrow \emptyset $ // \textit{Initialization}
  \WHILE{$i < N_p$}  
  \STATE $\bm{\alpha}^{(i)} \leftarrow Sampling(\mathcal{A})$;
  \STATE $\mathcal{\hat{L}}^{{(i)}}\leftarrow M(\bm{\alpha}^{(i)})$;
  \STATE $P \leftarrow P \cup \bm{\alpha}^{(i)}$, $\mathcal{\hat{L}} \leftarrow \mathcal{\hat{L}} \cup \mathcal{\hat{L}}^{{(i)}}$;
  \STATE $i\leftarrow{i+1}$;
  \ENDWHILE
  \WHILE{$t < T-1$}
      \STATE \textcolor{black}{$P^{'} \leftarrow Reproduction(P, p_c, p_m)$;}
      \STATE $\mathcal{\hat{L}}^{'}\leftarrow M(P^{'})$; // \textit{Evaluation}
      \STATE \textcolor{black}{$P,\mathcal{\hat{L}} \leftarrow Selection(P \cup P^{'},\mathcal{\hat{L}} \cup \mathcal{\hat{L}}^{'})$;}
      \STATE $t\leftarrow{t+1}$;
  \ENDWHILE
  \STATE $\bm{\alpha}^*\leftarrow$Select the best architecture in $P$ \textcolor{black}{based on $\mathcal{\hat{L}}$};
  \STATE $\bm{w}^*\leftarrow Retrain(\bm{\alpha}^*)$;
 \end{algorithmic}
\end{algorithm}

\subsubsection{Evolutionary Search} The framework of EGTAS is presented in Algorithm \ref{alg2}, which follows the basic evolutionary search. First, a surrogate model $M$ that fits the functional relationship between the architectural encoding and performance metric is constructed (See Algorithm \ref{alg1}). Then, we initialize a population \textcolor{black}{$P$} by randomly sampling a set of architectures and evaluate their performance metrics \textcolor{black}{$\mathcal{\hat{L}}$} by the surrogate model \textcolor{black}{$M$} \textcolor{black}{(see Steps 2-9 of Algorithm \ref{alg2}).} Note that these architectures are encoded as individuals in the population through the proposed encoding strategy. Next, we maintain a population to approach the optimal architecture through \textcolor{black}{\textit{Reproduction}, \textit{Evaluation}, and \textit{Selection}}. \textcolor{black}{\textit{Reproduction} operates on the parent population \textcolor{black}{$P$} to generate the offspring population \textcolor{black}{$P^{'}$}. Due to the proposed general encoding strategy, classical reproduction operators like the two-point crossover and polynomial mutation can be readily employed \cite{jin2021data}. All offspring undergo rapid performance evaluation using a cheap surrogate model. For \textit{Selection}, the top half of individuals based on performance metrics $\mathcal{\hat{L}} \cup \mathcal{\hat{L}}^{'}$ are chosen from the merged population $P \cup P^{'}$ of parents and offspring to form the new population $P$ (see Step 13 of Algorithm \ref{alg2}).} Finally, the best architecture \textcolor{black}{$\bm{\alpha}^*$} in the last population is retrained.

\begin{figure*}[t]
\centering
\includegraphics[width=0.7\textwidth]{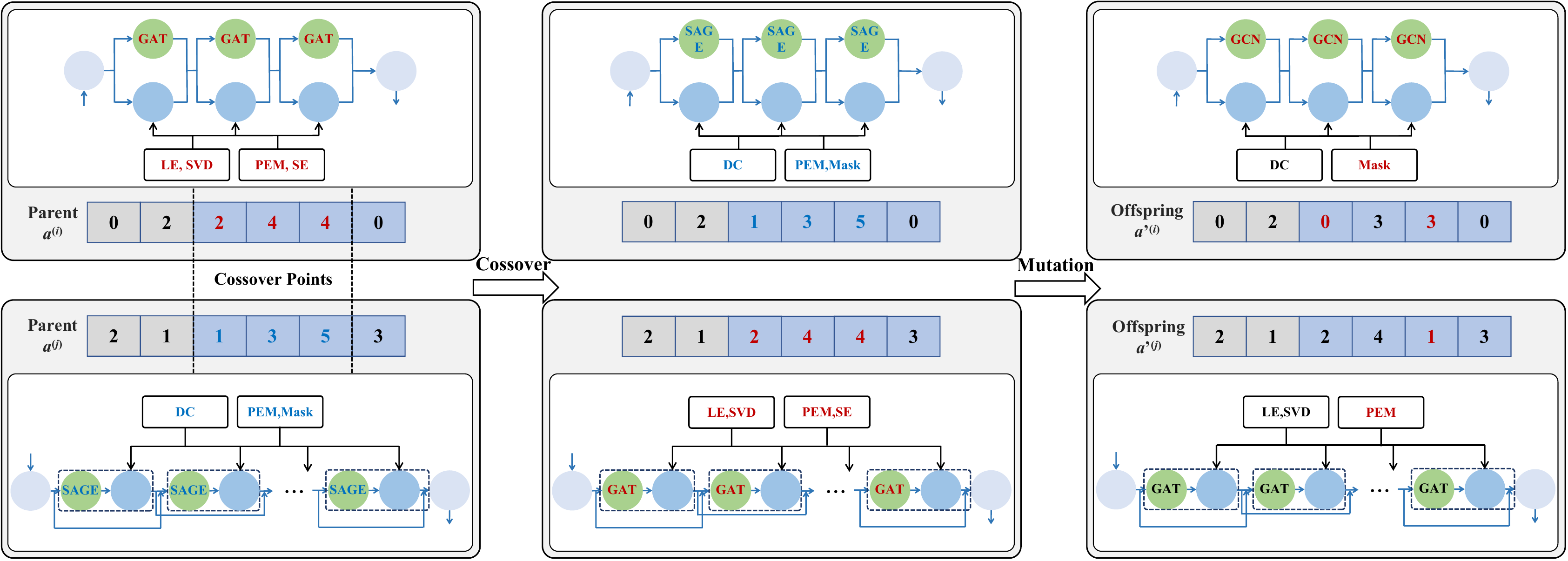}
\caption{\textcolor{black}{An illustrative example of a reproduction operator involving crossover and mutation.}} \label{reproduction}
\end{figure*}

\textcolor{black}{Fig. \ref{reproduction} is an illustrative example of a reproduction operator involving crossover and mutation. Crossover recombines genetic fragments across any two individuals (called parents) in the parent population with probability $p_c$ to generate new individuals. Taking two-point crossover as an example, the integer vectors located between two randomly chosen crossover points are swapped. From an architectural perspective, the operations of two architectures are combined. This combination allows for the preservation of excellent operations from the parents in evolutionary search. Mutation perturbs each individual generated by the crossover to generate a new offspring. Each element in the integer vector may change near its original value with probability $p_m$. From the standpoint of architecture, new operations are selected from a candidate set to replace existing ones, thereby enhancing the diversity of architectures. In each generation, the reproduction operators, including crossover and mutation, are repeatedly applied to generate the entire offspring population.}

\subsubsection{\textcolor{black}{Complexity Analysis}} \textcolor{black}{EGTAS consists of surrogate model construction and evolutionary search, whose time complexity is mainly determined by the training of sampling architectures (see Step 2 of Algorithm \ref{alg1}) and performance evaluation (see Step 12 of Algorithm \ref{alg2}), respectively. Assuming the number of training epochs is $N_e$, the size of a training set is $|\mathcal{D}_{tra}|$, and the maximum number of parameters in sampling architectures is $p$, the time complexity of training $N_s$ architectures is $O(N_s\times N_e\times|\mathcal{D}_{tra}|\times p )$. Evolutionary search requires $N_p\times T$ performance evaluations using the cheap surrogate model, where $N_p$ and $T$ are accordingly the population size and number of iterations. Since the training of sampling architectures is much more expensive than performance evaluation, the overall time complexity of EGTAS can be considered as $O(N_s\times N_e\times|\mathcal{D}_{tra}|\times p )$.}

\begin{table}[t]
\scriptsize
\centering
\caption{Details of real-world graph datasets for the node classification task.}\label{nc_datasets}
\begin{tabular}{cccccc}
\toprule
Datasets & \#Nodes & \#Edges & \#Features & \#Classes & \textcolor{black}{Sample} \\
\midrule
Cora & 2,708 & 5,429 & 1,433 & 7 & \textcolor{black}{Full Batch} \\
Computers & 13,381 & 245,778 & 767 & 10 & \textcolor{black}{Full Batch} \\
PubMed & 19,717 & 44,338 & 500 & 3 & \textcolor{black}{Full Batch} \\
Wisconsin & 251 & 466 & 1,703 & 5 & \textcolor{black}{Full Batch} \\
Flickr & 89,250 & 899,756 & 500 & 7 & \textcolor{black}{Full Batch} \\
\textcolor{black}{ogbn-arxiv} & \textcolor{black}{169,343} & \textcolor{black}{1,166,243} & \textcolor{black}{128} & \textcolor{black}{40} & \textcolor{black}{Neighbor} \\
\textcolor{black}{ogbn-products} & \textcolor{black}{2,449,029} & \textcolor{black}{61,859,140} & \textcolor{black}{100} & \textcolor{black}{47} & \textcolor{black}{Neighbor} \\
\bottomrule
\end{tabular}
\end{table}

\section{Experiments}
\textcolor{black}{In this section, to verify the effectiveness of EGTAS, we first introduce a series of experimental studies conducted on benchmark and real-world graph datasets for both node-level and graph-level tasks.} Subsequently, ablation experiments show the rationality of the constructed search space and surrogate model. Finally, a critical parameter (Model Scale) is further analyzed to explore the over-smoothing problems. We run all experiments on a Linux platform with a GPU 2080Ti (Memory: 12GB, Cuda Version:11.3).

\subsection{Node Classification Task}
\subsubsection{Experimental Settings}
\paragraph{Datasets}
\textcolor{black}{The statistics of real-world graph datasets for the node classification task are presented in Table \ref{nc_datasets}, which covers different types of complex networks.} Cora and Pubmed describe citation networks, where each node is the paper, and each edge is the citations across papers. Computers describe co-purchase networks of Amazon. Wisconsin represents the hyperlink relationship across pages. Flickr represents common attributes across images. \textcolor{black}{ogbn-arxiv comprises a directed graph that depicts the citation network encompassing computer science arXiv papers. ogbn-products consists of an undirected and unweighted graph that represents the Amazon product co-purchasing network.} \textcolor{black}{We can directly download these datasets from PyTorch Geometric\footnote{\url{https://github.com/pyg-team/pytorch\_geometric}\label{pyg}} and ogb\footnote{\url{https://github.com/snap-stanford/ogb}\label{ogb}}.} For Cora, Computers, and PubMud, the nodes of each class are split into 60\%/20\%/20\% for $\mathcal{D}_{tra}$/$\mathcal{D}_{val}$/$\mathcal{D}_{test}$ \cite{10.1145/3485447.3512185}. For Wisconsin, we split the dataset with 48\%/32\%/20\% for $\mathcal{D}_{tra}$/$\mathcal{D}_{val}$/$\mathcal{D}_{test}$ \cite{Pei2020Geom,NEURIPS2020_58ae23d8}. For Flickr, nodes are split into 50\%/25\%/25\% for $\mathcal{D}_{tra}$/$\mathcal{D}_{val}$/$\mathcal{D}_{test}$ \cite{Zeng2020GraphSAINT}. \textcolor{black}{For ogbn-arxiv and ogbn-products, we use the official splits provided by ogb \cite{hu2020open}. Due to their extensive size, we adopt neighborhood sampling following GraphSAGE \cite{hamilton2017inductive}. Specifically, we sample 20 neighbors per layer for ogbn-arxiv and 15 for ogbn-products.} For a fair comparison, all dataset configurations are implemented via the publicly available codes\footnote{\url{https://github.com/LARS-research/F2GNN}\label{f2}}.

\begin{table}[t]
\centering
\caption{\textcolor{black}{Training hyper-parameters of EGTAS. $D$ is encoding dimension.}}\label{nc_hyperparameters}\color{black}
\begin{tabular}{ccc}
\toprule
Category & Hyper-parameters & Values \\
\midrule
        \multirow{8}{*}{\makecell{Surrogate Model \\ Construction}} & Number of Samples & 30 \\
        & Maximum Steps & 1e+4 \\
        & Warm-up Steps & 4e+3 \\
        & Attention Dropout & 0.5 \\
        & FFN Dropout & 0.1 \\
        & GNN Dropout & 0.5 \\
        & Learning Rate & 2e-4 \\
        & Weight Decay & 1e-5 \\
        & Batch Size & \{32,128\} \\
\midrule
        \multirow{3}{*}{\makecell{Evolutionary \\ Search}} & Population Size  & 20 \\
        & Crossover Probability & 0.7\\
        & Mutation Probability & $1/D$ \\
\midrule
       \multirow{2}{*}{Retrain} & Maximum Steps & \{1e+6,3e+6\} \\
        & Warm-up Steps & 4e+4 \\
\bottomrule
\end{tabular}
\end{table}

\paragraph{Baselines}
For the node classification task, we provide three types of state-of-the-art baselines: Manual-GNN, Manual-GT, and GNAS. Manual-GNN and Manual-GT contain the popular human-designed GNNs and graph Transformers including GCN \cite{kipf2017semisupervised}, SAGE \cite{hamilton2017inductive}, GAT \cite{veličković2018graph}, GATv2 \cite{brody2022how}, GIN \cite{xu2018how}, Graphormer \cite{ying2021do}, and GMT \cite{10.1145/3477495.3532031}. In addition, we employ three topology designs mentioned in this paper as the baseline, Res-GAT, JK-GAT, and GCNII-GAT, where GAT is adopted as the aggregation operation. \textcolor{black}{GNAS baselines include SNAG \cite{zhao2020simplifying}, SANE \cite{9458743}, F$^2$GNN \cite{10.1145/3485447.3512185}, and GAUSS \cite{guan2022large}.} SNAG is an advanced RL-based method to automate the search for GNN architectures. SANE and F$^2$GNN are state-of-the-art differentiable methods to search aggregate operations and topologies, respectively. \textcolor{black}{GAUSS is a state-of-the-art GNAS method specifically designed for large-scale graphs.}

\paragraph{Implementation Details} We implement all baselines by three open libraries: PyTorch Geometric (v. 2.1.0), Transformers (v. 4.17.0), and Geatpy (v. 2.7.0). For all baselines, we report the final results based on ten independent runs. For the Manual-GNN and GNAS baselines, the number of backbones is set to four. All Manual-GNN baselines are implemented by the PyTorch Geometric\textsuperscript{\ref {pyg}}. For Res-GAT\textsuperscript{\ref {f2}}, JK-GAT\textsuperscript{\ref {f2}}, GCNII-GAT\textsuperscript{\ref {f2}}, Graphormer\footnote{\url{https://github.com/qwerfdsaplking/Graph-Trans}\label{gtt}}, GMT\textsuperscript{\ref {gtt}}, SNAG\footnote{\url{https://github.com/LARS-research/SNAG}}, SANE\footnote{\url{https://github.com/LARS-research/SANE}}, and F$^2$GNN\textsuperscript{\ref {f2}}, the codes by the original authors are adopted. \textcolor{black}{The experimental results of GAUSS on large-scale datasets are reported in \cite{guan2022large}.} We implement the evolutionary search of the proposed EGTAS via Geatpy\footnote{\url{https://github.com/geatpy-dev/geatpy}} with the default settings, \textcolor{black}{applying a \textit{strengthened elitist genetic algorithm} that incorporates two-point crossover, polynomial mutation, and elitist selection. Based on empirical evidence, the crossover probability and mutation probability are set to 0.7 and $1/D$, respectively, where $D$ denotes the encoding dimension. Since the search space contains six operations, $D$ is 6 in this paper.} The population size is set to 20. For surrogate model construction in EGTAS, training maximum steps and the number of samples are set to 1e+4 and 30, respectively, for the sake of computational resources. The Adam optimizer with the peak learning rate of 2e-4 is adopted. \textcolor{black}{For a fair comparison, we retrain the architectures searched by each GNAS baseline. For the large-scale ogbn-products dataset, we use a maximum step of 3e+6 and a batch size of 32, whereas for all other datasets, the settings are 1e+6 and 128 respectively. The training hyperparameters of EGTAS are summarized in Tables \ref{nc_hyperparameters}.} A more detailed introduction to the hyper-parameter of baselines can be found in \cite{10.1145/3485447.3512185,min2022transformer}. All the hyper-parameters are tuned via the publicly available codes\textsuperscript{\ref {f2},\ref {gtt}}.

\begin{table*}[t]
\begin{center}
\caption{Experimental results of all baselines on the node-level classification task in terms of \#Acc (\%$\uparrow$). N/A denotes the methods that have not provided test results for the specified graph dataset.}
\label{node_classification}
\footnotesize
\begin{tabular}{ccccccccc}
\toprule
Category & Baselines & Cora & Computers & PubMed & Wisconsin & Flickr & \textcolor{black}{ogbn-arxiv} & \textcolor{black}{ogbn-products} \\
\midrule
        \multirow{8}{*}{Manual-GNN} & GCN & 86.19(0.83) & 80.74(1.96) & 85.38(0.15) & 51.87(6.22) & 51.30(0.07) & \textcolor{black}{71.60(0.15)} & \textcolor{black}{78.97(0.33)}  \\
        & SAGE & 85.68(0.61) & 90.52(0.42) & 88.23(0.28) &  60.39(10.77) &  53.07(0.50) & \textcolor{black}{71.49(0.27)} & \textcolor{black}{78.70(0.36)}  \\
        & GIN & 85.81(1.89) & 72.96(4.87) & 87.14(1.35) & 55.26(5.84) & 48.76(0.43) & \textcolor{black}{70.36(0.34)} & \textcolor{black}{79.07(0.52)}  \\
        & GAT & 86.16(0.55) &  89.08(0.43) & 85.73(0.34) &  45.29(5.65) &  50.34(2.68) & \textcolor{black}{71.54(0.30)} & \textcolor{black}{77.23(2.37)}  \\
        & GATv2 & 86.97(0.60) & 88.25(7.71) & 90.27(0.32) & 46.47(4.39) & 50.48(0.63) & \textcolor{black}{71.59(0.16)} & \textcolor{black}{78.46(2.45)}  \\
        & Res-GAT & 84.66(0.92) & 90.84(0.49) & 87.56(0.44) & 48.82(3.77) & 53.63(0.24) & \textcolor{black}{71.63(0.94)} &  \textcolor{black}{79.84(0.34)} \\
        & JK-GAT & 86.55(0.46) & 91.80(0.23) & 89.71(0.16) & 84.51(5.58) & 53.02(0.29) & \textcolor{black}{71.75(0.14)} & \textcolor{black}{80.31(1.02)} \\
        & GCNII-GAT & 85.40(1.06) & 91.91(0.11) & 88.44(0.25) &  55.29(6.25) &  53.03(0.29) & \textcolor{black}{71.53(0.31)} & \textcolor{black}{78.34(0.58)} \\
\midrule
        \multirow{2}{*}{Manual-GT} & Graphormer & 88.95(0.53) & 90.36(0.90) & 84.23(0.17) & 42.00(2.22) & 52.68(0.29) & \textcolor{black}{70.98(0.67)} & \textcolor{black}{79.34(0.25)} \\
        & GMT & 87.84(0.89) & 90.62(0.51) & 83.87(0.74) & 50.00(3.02) & 53.54(0.21) & \textcolor{black}{71.82(0.14)} & \textcolor{black}{79.51(0.36)}  \\
\midrule
       GNAS-RL & SNAG & 84.99(1.04) & 85.98(0.72) & 87.93(0.16) &  43.92(4.65) & 53.50(0.31) & \textcolor{black}{71.29(0.31)} & \textcolor{black}{79.47(0.93)} \\
       GNAS-Gradient & SANE & 86.40(0.38) & 91.02(0.21) & 89.34(0.31) & 86.47(3.09) & 53.92(0.14) & \textcolor{black}{71.77(0.51)}&  \textcolor{black}{80.66(0.75)} \\
       GNAS-Gradient & F$^2$GNN & 87.42(0.42) & 91.42(0.26) & 89.79(0.20) & 88.24(3.72) & 53.96(0.20) & \textcolor{black}{71.93(0.30)}  & \textcolor{black}{80.91(0.76)} \\
       \textcolor{black}{GNAS-Gradient} & \textcolor{black}{GAUSS} & \textcolor{black}{N/A} & \textcolor{black}{N/A} & \textcolor{black}{N/A} & \textcolor{black}{N/A} &\textcolor{black}{N/A} & \textcolor{black}{72.35(0.21)} & \textcolor{black}{81.26(0.36)}  \\
\midrule
       GNAS-EA+NP & EGTAS & \textbf{89.14(0.80)} & \textbf{92.53(0.66)} & \textbf{90.75(0.31)} & \textbf{89.12(3.10)} & \textbf{54.65(0.62)} & \textcolor{black}{\textbf{72.75(0.61)}} & \textcolor{black}{\textbf{82.15(0.37)}}  \\      
\bottomrule
\end{tabular}
\end{center}
\end{table*}

\begin{figure*}[t]
\centering
\includegraphics[width=0.8\textwidth]{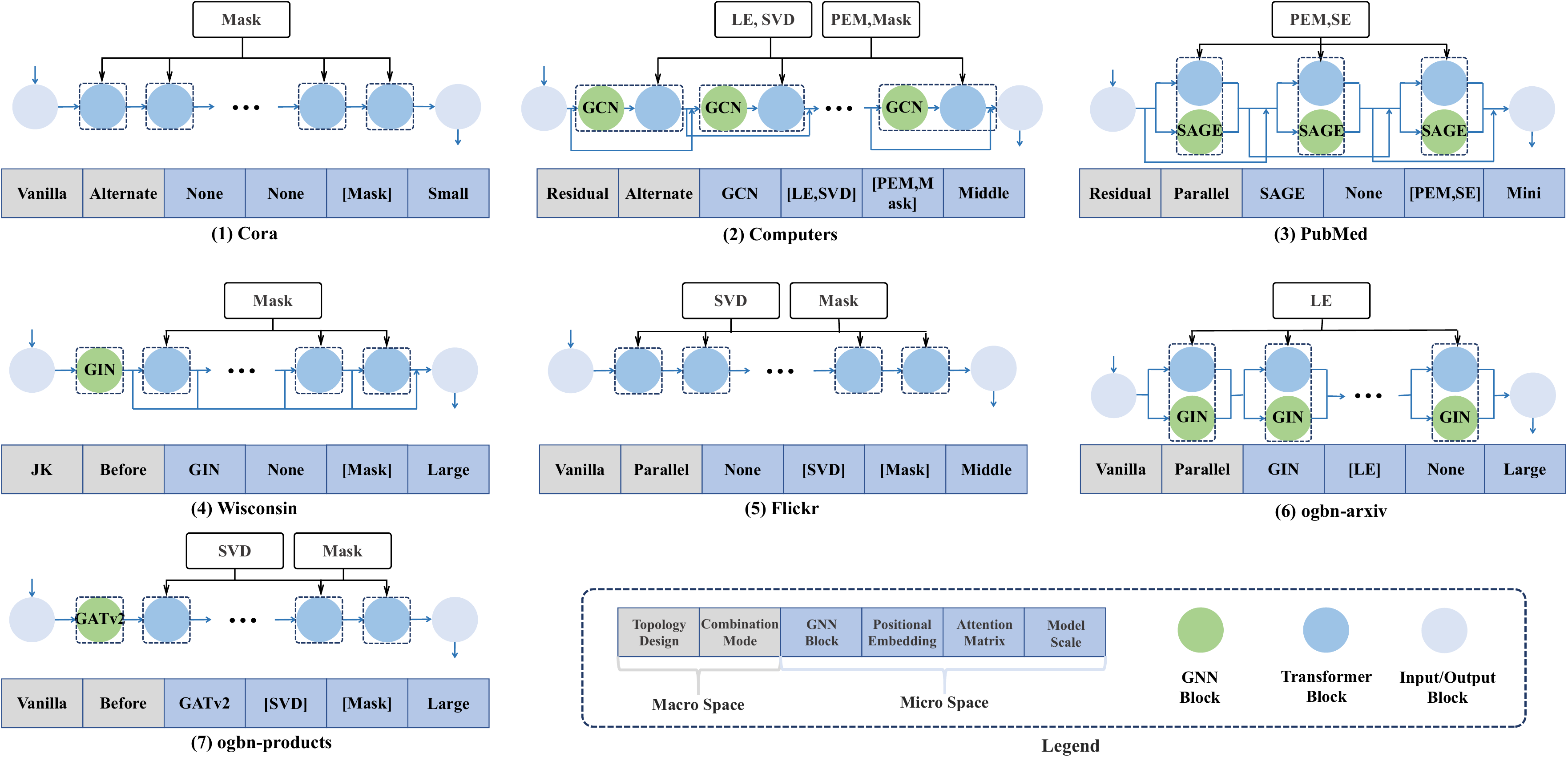}
\caption{\textcolor{black}{Visualization of architectures searched by EGTAS on all datasets for the node classification task.}} \label{sa}
\end{figure*}

\begin{figure}[htbp]
\centering
\includegraphics[width=0.45\textwidth]{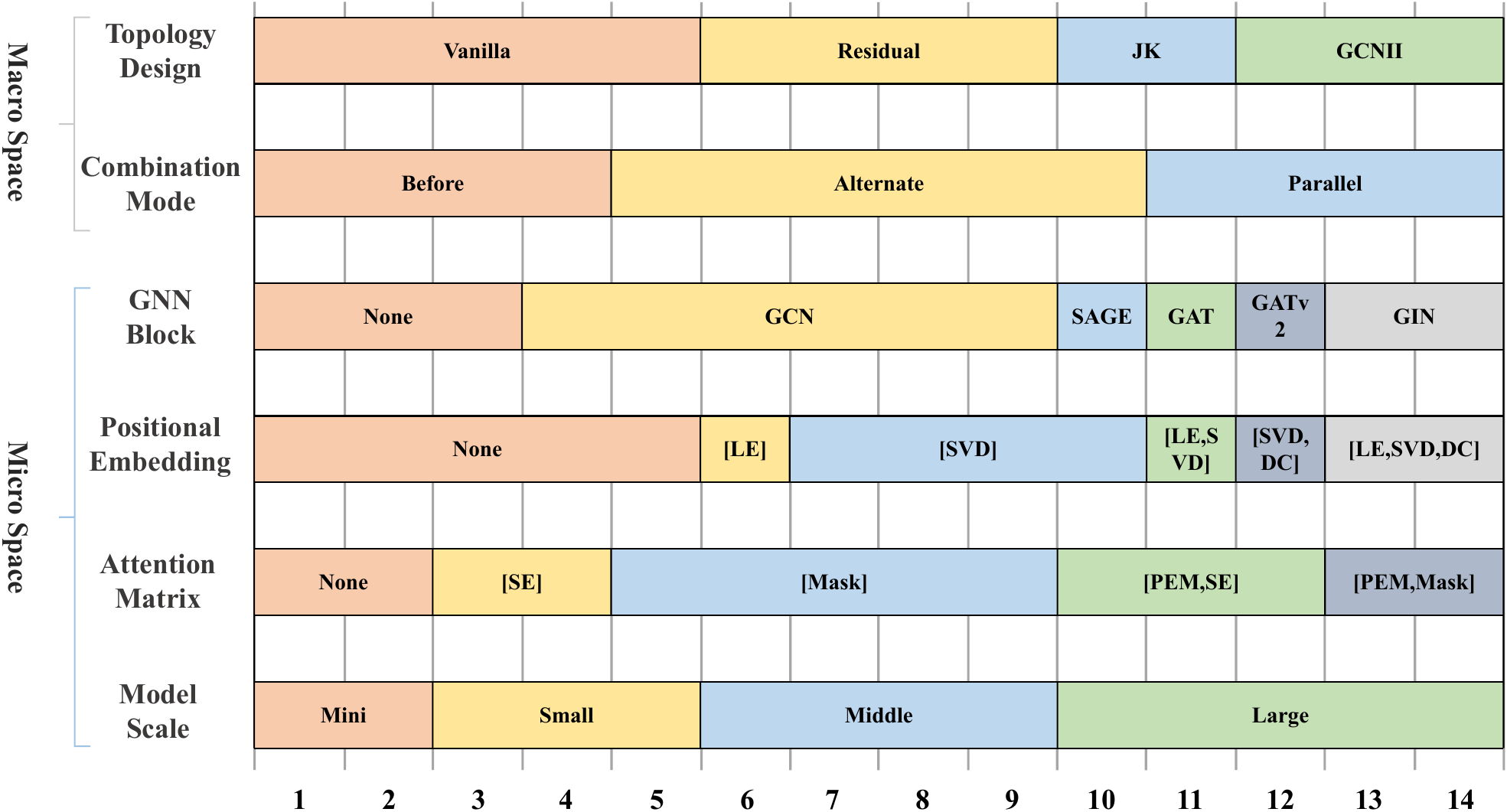}
\caption{\textcolor{black}{Statistics for top 10\% architectures in the last population on all datasets for the node classification task.}} \label{countsa}
\end{figure}

\subsubsection{Experimental Results}
\paragraph{Performance Comparisons}
Table \ref{node_classification} lists the experimental results of all baselines on the node classification task in terms of \#Acc. We report the average value and the standard deviation. The best result is marked in bold. From Table \ref{node_classification}, we observe that the model searched by EGTAS rival the best manually designed ones (GNN and Graph Transformer), which shows that EGTAS can design a graph Transformer model with strong representation ability. Furthermore, by adaptively searching topologies and graph-aware strategies, EGTAS achieves considerable performance gains over GNAS baselines on all datasets. This phenomenon demonstrates the necessity of exploring graph Transformer architectures from both macro and micro levels.

\paragraph{Searched Architectures}

Fig. \ref{sa} visualizes the optimal architectures searched by EGTAS in all cases. The searched macro-ops and micro-ops are significantly different, which further illustrates the importance of studying graph Transformer architecture search. That is, it is necessary to customize exclusive architectures for different graph datasets. 
\textcolor{black}{Compared to small-scale graphs (Cora, Computers, PubMed), the architectures searched on large-scale graphs (Flickr, ogbn-arxiv, ogbn-products) have larger model scale, which is in line with the general intuition of designing deep models, i.e. more hidden layers to learn deeper feature representations. For graph-aware strategies, we find that not all positional embedding and attention matrix operations are adopted in the searched architectures, which indicates that simply combining graph-aware strategies manually is not optimal. We collect the top 10\% architectures in the last population obtained by EGTAS on all datasets and count the number of macro-ops and micro-ops as shown in Fig. \ref{countsa}. The results show a mostly uniform distribution of operations, with no single operation standing out as highly probable. This phenomenon illustrates that no single architecture performs best on all data sets, thus highlighting the necessity of finding the optimal combination of different operations.}

\paragraph{Time Cost}
We further explore the time comparison of EGTAS with hand-designed graph Transformer models using the Flickr dataset as an example. The default Graphormer and GMT are trained for 1e+6 maximum steps, which cost about 55 hours and 33 hours on a single GPU, respectively. The total search cost consumed by the EGTAS can be divided into two parts: surrogate model construction and evolutionary search. In surrogate model construction, 30 sampled architectures are independently trained for 1e+4 maximum steps, which costs about 17 hours on a single GPU. In the evolutionary search phase, each individual in the population is evaluated by the cheap surrogate model, which takes only a few seconds on a CPU. Therefore, the total search cost of EGTAS is mainly determined by the surrogate model construction, which is less than the training cost of a hand-designed model.

\subsection{Graph Classification Task}

\begin{table*}[htbp]
\footnotesize
\centering
\caption{Details of real-world graph datasets for the graph classification task.}\label{gc_datasets}
\begin{tabular}{cccccccc}
\toprule
Datasets & \#Graphs & \#Avg. Nodes & \#Avg. Edges & \#Node Features & \#Edge Features & \#Classes & \textcolor{black}{\#Metric} \\
\midrule
DHFR\_MD & 393 & 23.87 & 283.01 & 7 & 5 & 2 &  \textcolor{black}{\#Acc} \\
ogbg-molbbbp & 2,039 & 34.09 & 36.86 & 9 & 3 & 2 & \textcolor{black}{\#AUC} \\
ogbg-molbace & 1,513 & 25.51 & 27.47 & 9 & 3 & 2 & \textcolor{black}{\#AUC} \\
ogbg-molhiv & 41,127 & 24.06 & 25.95 & 9 & 3 & 2 & \textcolor{black}{\#AUC} \\
\bottomrule
\end{tabular}
\end{table*}

The statistics of datasets for the graph classification task are introduced in Table \ref{gc_datasets}. DHFR\_MD\footnote{\url{https://chrsmrrs.github.io/datasets/}\label{tud}} is a collection of compounds derived from the study of dihydrofolate reductase inhibitors. OGBG-MolBBBP, OGBG-MolBACE, and OGBG-MolHIV from ogb \cite{hu2020open} are molecular property prediction datasets where each graph is a molecule. In a molecule, nodes are atoms and edges are chemical bonds.

\begin{table*}[htbp]
\begin{center}
\caption{Experimental results of all baselines on the graph-level classification task in terms of \#Acc (\%$\uparrow$) or \#AUC (\%$\uparrow$).}
\label{graph_classification}
\footnotesize
\begin{tabular}{cccccc|c}
\toprule
Category & Baselines & DHFR\_MD (\#Acc) & ogbg-molbbbp (\#AUC) & ogbg-molbace (\#AUC) & ogbg-molhiv (\#AUC)  & AR \\
\midrule
    Manual-GNN & GIN & 62.88(8.26) & 63.37(1.81) & 70.42(4.78) & 71.11(2.57) & 4.0 \\
    Manual-GT & Graphormer & 64.88(7.58) & 66.52(0.74) & 76.42(1.67) & 71.89(2.66) & 3.0 \\
\midrule
   GNAS-EA+ST & AutoGT & 68.22(5.02) & 67.29(1.46) & 76.70(1.42) & 74.95(1.02) & 2.0 \\
\midrule
   GNAS-EA+NP & EGTAS & \textbf{69.12(5.64)}  & \textbf{70.33(1.13)} & \textbf{79.19(1.37)} & \textbf{79.81(1.17)} & 1.0 \\
\bottomrule
\end{tabular}
\end{center}
\end{table*}

Three state-of-the-art baselines are considered for the graph classification task: GIN \cite{xu2018how}, Graphormer \cite{ying2021do}, and AutoGT \cite{zhang2023autogt}. GIN and Graphormer are the popular human-designed GNN and Transformer models, respectively. AutoGT is a state-of-the-art GNAS approach to search for graph Transformer architectures for graph classification tasks, which follows the basic framework of evolutionary search and supernet training (ST). The algorithm configuration of the proposed EGTAS is consistent with Section V.A. We employ ten-fold cross-validation for all baselines. For a fair comparison, all hyper-parameter and training strategy configurations are kept the same as the original paper of AutoGT \cite{zhang2023autogt}.

Table \ref{graph_classification} lists the experimental results (average value and standard deviation) of all baselines on the graph classification task in terms of \#Acc or \#AUC. AR refers to the average rank of the baseline on each dataset. GNAS baselines (EGTAS and AutoGT) outperform handcrafted baselines (GIN and Graphormer) on all datasets. Since the manually designed architecture cannot adapt to different datasets, it is not as effective as the automatic GNAS method. EGTAS significantly outperforms AutoGT when the number of training data is large, such as ogbg-molhiv. When the number of training data is small, such as DHFR\_MD, the improvement of EGTAS is not obvious. This illustrates the necessity of jointly designing topologies and graph-aware strategies for datasets with different scales.

\subsection{\textcolor{black}{Benchmarking GNNs}}

\begin{table*}[htbp]
\footnotesize
\centering
\caption{\textcolor{black}{Details of benchmark graph datasets used in this study.}}\label{benchmark_datasets}\color{black}
\begin{tabular}{cccccccc}
\toprule
Datasets & \#Graphs & \#Avg. Nodes & \#Avg. Edges & \#Node Features & \#Classes & \#Task Types & \#Metric \\
\midrule
ZINC & 12,000 & 23.2 & 49.8 & 28 & 1 & Regression & \#MAE \\
PATTERN & 14,000 & 118.9 & 6,098.9 & 3 & 2 & Node Cla. & \#Acc \\
CLUSTER & 12,000 &  117.2 & 4,303.9 & 7 & 6 & Node Cla. & \#Acc \\
MNIST & 70,000 & 70.6 & 564.5 & 3 & 10 & Graph Cla. & \#Acc \\
CIFAR10 & 60,000 & 117.6 & 941.2 & 5 & 10 & Graph Cla. & \#Acc \\
\bottomrule
\end{tabular}
\end{table*}

\textcolor{black}{We evaluate the performance of EGTAS on five benchmark graph datasets from Benchmarking GNNs \cite{JMLR:v24:22-0567}, namely ZINC, PATTERN, CLUSTER, MNIST, and CIFAR10. These datasets encompass a wide range of graph tasks, including regression, node classification, and graph classification, highlighting the versatility and generality of our study. The detailed statistics of these datasets are presented in Table \ref{benchmark_datasets}. ZINC includes 12K molecular graphs from the ZINC database. Nodes represent heavy atoms and edges represent bonds. MNIST and CIFAR10 are graph datasets derived from image classification datasets. Each image is converted into an 8 nearest-neighbor graph of SLIC superpixels. PATTERN and CLUSTER are synthetic datasets generated with Stochastic Block Models that model communities in social networks. The task is to classify which nodes belong to the same community. For a fair comparison, we adopt the standard splits used in benchmarking GNNs \cite{JMLR:v24:22-0567}.}

\begin{table*}
\footnotesize
\centering
\caption{\textcolor{black}{Training hyper-parameters of EGTAS for benchmark graph datasets. $D$ is encoding dimension.}}\label{bench_hyperparameters}\color{black}
\begin{tabular}{ccccccc}
\toprule
Category & Hyper-parameters & ZINC & PATTERN & CLUSTER & MNIST & CIFAR10 \\
\midrule
        \multirow{8}{*}{\makecell{Surrogate Model \\ Construction}} & Number of Samples & 20 & 20 & 20 & 20 & 20 \\
        & Maximum Epochs & 50 & 10 & 10 & 10 & 10 \\
        & Warm-up Epochs & 10 & 5 & 5 & 5 & 5 \\
        & Attention Dropout & 0.5 & 0.5 & 0.5 & 0.5 & 0.5 \\
        & FFN Dropout & 0.1 & 0.1 & 0.1 & 0.1 & 0.1 \\
        & GNN Dropout & 0.5 & 0.5 & 0.5 & 0.5 & 0.5 \\
        & Learning Rate & 1e-3 & 5e-4 & 5e-4 & 1e-3 & 1e-3 \\
        & Weight Decay & 1e-5 & 1e-5 & 1e-5 & 1e-5 & 1e-5 \\
        & Batch Size & 32 & 32 & 16 & 16 & 16 \\
\midrule
        \multirow{3}{*}{\makecell{Evolutionary \\ Search}} & Population Size  & 20 & 20 & 20 & 20 & 20 \\
        & Crossover Probability & 0.7 & 0.7 & 0.7 & 0.7 & 0.7 \\
        & Mutation Probability & $1/D$ & $1/D$ & $1/D$ & $1/D$ & $1/D$ \\
\midrule
       \multirow{2}{*}{Retrain} & Maximum Epochs & 2000 & 100 & 100 & 100 & 100 \\
        & Warm-up Epochs & 50 & 5 & 5 & 5 & 5 \\
\bottomrule
\end{tabular}
\end{table*}

\textcolor{black}{We compare EGTAS with popular manual baselines: GCN \cite{kipf2017semisupervised}, GAT \cite{veličković2018graph}, GIN \cite{xu2018how}, GatedGCN \cite{bresson2017residual}, PNA \cite{corso2020principal}, DGN \cite{beaini2021directional}, CIN \cite{bodnar2021weisfeiler}, CRaWl \cite{toenshoff2021graph}, GIN-AK+ \cite{zhao2021stars}, SAN \cite{kreuzer2021rethinking}, Graphormer \cite{ying2021do}, K-Subtree SAT \cite{chen2022structure}, EGT \cite{hussain2021edge}, and GPS \cite{rampavsek2022recipe}. These baselines cover a variety of well-known message-passing neural networks and graph transformer models, which are reported in \cite{rampavsek2022recipe}. All baseline results are based on ten independent runs. The algorithm configuration for EGTAS aligns with the description in Section V.A. The training hyperparameters for benchmark graph datasets are presented in Tables \ref{bench_hyperparameters}.}

\begin{table*}[htbp]
\begin{center}
\footnotesize
\caption{\textcolor{black}{Experimental results of all baselines on the benchmark graph datasets in terms of \#MAE (\%$\downarrow$) or \#Acc (\%$\uparrow$). N/A denotes the methods that have not provided test results for the specified graph dataset.}}
\label{benchmarking_results}\color{black}
\begin{tabular}{cccccc}
\toprule
Baselines & ZINC (\#MAE) & PATTERN (\#Acc) & CLUSTER (\#Acc) & MNIST (\#Acc) & CIFAR10 (\#Acc) \\
\midrule
 GCN & 0.367(0.011) & 71.892(0.334) & 68.498(0.976) & 90.705(0.218) & 55.710(0.381)  \\
 GAT & 0.384(0.007) & 78.271(0.186) & 70.587(0.447) & 95.535(0.205) & 64.223(0.455) \\
 GIN & 0.526(0.051) & 85.387(0.136) & 64.716(1.553) & 96.485(0.252) & 55.255(1.527) \\
 GatedGCN & 0.282(0.015) & 85.568(0.088) & 73.840(0.326) & 97.340(0.143) & 67.312(0.311) \\
 PNA & 0.188(0.004) & N/A & N/A & 97.94(0.12) &  70.35( 0.63) \\
 DGN & 0.168(0.003) & 86.680(0.034) & N/A & N/A & 72.838(0.417) \\
 CIN & 0.079(0.006) & N/A & N/A & N/A & N/A \\
 CRaWl & 0.085(0.004) & N/A & N/A & 97.944(0.050) & 69.013(0.259) \\
 GIN-AK+ & 0.080(0.001) & \textbf{86.850(0.057)} & N/A & N/A & 72.19(0.13) \\
\midrule
 SAN & 0.139(0.006) & 86.581(0.037) & 76.691(0.65) & N/A & N/A \\
 Graphormer & 0.122(0.006) & N/A & N/A & N/A & N/A \\
 K-Subtree SAT & 0.094(0.008) & 86.848(0.037) & 77.856(0.104) & N/A & N/A \\
 EGT & 0.108(0.009) & 86.821(0.020) & 79.232(0.348) & 98.173(0.087) & 68.702(0.409) \\
 GPS & \textbf{0.070(0.004)} & 86.685(0.059) & 78.016(0.180) & 98.051(0.126) & 72.298(0.356) \\
\midrule
 EGTAS & 0.075(0.073) & 86.742(0.053) & \textbf{79.236(0.215)} & \textbf{98.372(0.261)} & \textbf{74.308(0.575)} \\
\bottomrule
\end{tabular}
\end{center}
\end{table*}

\textcolor{black}{Table \ref{benchmarking_results} lists the experimental results (average value and standard deviation) of all baselines on the benchmark graph datasets. EGTAS outperforms state-of-the-art handcrafted baselines on three datasets, thus highlighting the ability to perform well on a variety of synthetic graph tasks. This once again demonstrates the importance of finding graph Transformer architectures to handle different datasets.}

\subsection{Ablation Studies}

\begin{table}[t]
\begin{center}
\caption{The ablation studies on the necessity of joint search micro and macro operations.}
\label{MA-MI}
\footnotesize
\begin{tabular}{cccc}
\toprule
Datasets & MA-EGTAS & MI-EGTAS & EGTAS \\
\midrule
        Cora & 81.72(1.07)$-$ & 80.41(3.99)$-$ & \textbf{89.14(0.80)} \\
        Computers & 90.47(0.44)$-$ & 91.30(0.36)$-$ & \textbf{92.53(0.66)} \\
        PubMed & 83.64(1.20)$-$ & 84.48(2.26)$-$ & \textbf{90.75(0.31)} \\
        Wisconsin & 70.56(5.12)$-$ & 86.12(6.46)$-$ & \textbf{89.12(3.10)} \\
        Flickr & 52.99(0.42)$-$ & 53.56(0.29)$-$ & \textbf{54.65(0.62)} \\
        \textcolor{black}{ogbn-arxiv} & \textcolor{black}{71.71(1.35)$-$} & \textcolor{black}{72.39(0.98)$-$} & \textcolor{black}{\textbf{72.75(0.61)}} \\
        \textcolor{black}{ogbn-products} & \textcolor{black}{80.05(0.65)$-$} & \textcolor{black}{81.53(2.76)}$-$ & \textcolor{black}{\textbf{82.15(0.37)}} \\
\midrule
$+/\approx/-$ & \textcolor{black}{$0/0/7$} & \textcolor{black}{$0/0/7$} & $-$ \\
\bottomrule
\end{tabular}
\end{center}
\end{table}

We present a series of studies on the node classification task to explore the effectiveness of the components of EGTAS. Two variants, MA-EGTAS and MI-EGTAS, are constructed to verify the necessity of joint search micro and macro operations. During the evolutionary search phase, MA-EGTAS only explores the macro search space and the components of the micro search space are randomly set as a fixed operation. Similarly, MI-EGTAS only explores the macro search space. Table \ref{MA-MI} records the experimental results (average value and the standard deviation) of EGTAS and its variants on the node classification task over 10 runs. Based on the Wilcoxon rank sum test with a 0.95 confidence level, symbols "$+/\approx/-$" indicate that the accuracy of the baseline is significantly better/similar/worse than that of EGTAS. \textcolor{black}{As can be seen from Table \ref{MA-MI}, EGTAS exhibits excellent performance in all cases in terms of \#Acc. Merely exploring either the macro space or micro space does not yield comparable results. The performance gain reveals the necessity of searching graph Transformer architectures at both macro and micro levels.}

\begin{table*}[t]
\caption{Comparison of six surrogate models on all datasets for the node classification task over ten runs.}
    \centering
    \label{SM}
    \resizebox{0.85\textwidth}{!}{
    \begin{tabular}{cccccccc}
    \toprule
       Datasets & Metrics & DT & RF & AdaBoost & Bagging & GP & ExtraTree \\
    \midrule
       \multirow{2}{*}{Cora} & KTau ($\uparrow$) &
7.33e-01(6.11e-01) &
6.67e-01(4.47e-01) &
8.00e-01(4.27e-01) &
8.67e-01(4.00e-01) &
\textbf{9.33e-01(2.00e-01)} &
8.00e-01(3.06e-01) \\
       & MSE ($\downarrow$) & \textbf{1.12e-04(1.02e-04)} &
1.40e-02(8.98e-03) &
1.58e-04(3.21e-04) &
3.18e-04(4.17e-04) &
1.07e-02(2.78e-02) &
1.81e-01(8.30e-02)  \\
       \multirow{2}{*}{Computers} & KTau ($\uparrow$) & 4.00e-01(9.17e-01) &
4.00e-01(9.17e-01) &
6.00e-01(8.00e-01) &
4.00e-01(9.17e-01) &
\textbf{8.00e-01(6.00e-01)} &
2.00e-01(9.80e-01) \\
       & MSE ($\downarrow$) & 5.41e-06(5.94e-06) &
4.22e-02(2.05e-02) &
\textbf{4.66e-06(4.11e-06)} &
7.36e-04(2.19e-03) &
3.77e-02(1.13e-01) &
4.90e-01(2.27e-01) \\
       \multirow{2}{*}{PubMed} & KTau ($\uparrow$) & 4.86e-01(2.30e-01) &
2.48e-01(4.07e-01) &
3.90e-01(3.39e-01) &
5.43e-01(2.16e-01) &
2.57e-01(3.58e-01) &
\textbf{8.01e-01(4.86e-01)} \\
       & MSE ($\downarrow$) & 3.89e-04(5.94e-04) &
4.53e-02(3.75e-02) &
\textbf{1.38e-04(4.09e-04)} &
1.75e-02(2.05e-02) &
1.30e-01(1.44e-01) &
1.15e-01(1.57e-01) \\
       \multirow{2}{*}{Wisconsin} & KTau ($\uparrow$) & \textbf{8.00e-01(3.06e-01)} &
3.33e-01(5.96e-01) &
6.00e-01(3.27e-01) &
7.33e-01(3.27e-01) &
4.00e-01(5.54e-01) &
0.00e+00(3.33e-01) \\
       & MSE ($\downarrow$) & 1.19e-03(7.99e-04) &
3.04e-03(2.77e-03) &
\textbf{6.94e-04(6.38e-04)} &
7.07e-04(6.47e-04) &
5.37e-03(1.13e-02) &
1.47e-01(6.94e-02) \\
       \multirow{2}{*}{Flickr} & KTau ($\uparrow$) & \textbf{8.48e-01(1.36e-01)} &
5.05e-01(2.29e-01) &
6.76e-01(1.96e-01) &
7.62e-01(1.30e-01) &
7.52e-01(2.14e-01) &
4.57e-01(2.76e-01) \\
       & MSE ($\downarrow$) & \textbf{9.75e-06(1.06e-05)} &
3.05e-03(2.57e-03) &
2.57e-05(1.36e-05) &
3.89e-04(7.41e-04) &
2.55e-05(2.31e-05) &
5.03e-02(2.59e-02) \\
       \multirow{2}{*}{\textcolor{black}{ogbn-arxiv}} & \textcolor{black}{KTau ($\uparrow$)} & 
\textcolor{black}{8.67e-01(2.21e-01)} &
\textcolor{black}{5.33e-01(4.52e-01)} &
\textcolor{black}{5.67e-01(5.39e-01)} &
\textcolor{black}{2.67e-01(4.90e-01)} &
\textcolor{black}{\textbf{9.67e-01(1.00e-01)}} &
\textcolor{black}{6.33e-01(3.79e-01)}
 \\
       & \textcolor{black}{MSE ($\downarrow$)} & \textcolor{black}{1.40e-02(1.71e-02)} &
\textcolor{black}{1.39e-02(8.53e-03)} &
\textcolor{black}{1.64e-02(7.29e-03)} &
\textcolor{black}{2.65e-02(3.19e-02)} &
\textcolor{black}{\textbf{3.31e-03(9.93e-03)}} &
\textcolor{black}{4.96e-02(6.14e-02)}
 \\
       \multirow{2}{*}{\textcolor{black}{ogbn-products}} & \textcolor{black}{KTau ($\uparrow$)} & \textcolor{black}{6.00e-01(6.11e-01)} &
\textcolor{black}{6.67e-01(4.47e-01)} &
\textcolor{black}{4.67e-01(5.81e-01)} &
\textcolor{black}{5.33e-01(5.21e-01)} &
\textcolor{black}{\textbf{8.00e-01(3.06e-01)}} &
\textcolor{black}{7.33e-01(3.27e-01)}
 \\
       & \textcolor{black}{MSE ($\downarrow$)} & \textcolor{black}{1.18e-01(1.96e-01)} &
\textcolor{black}{\textbf{4.56e-02(5.49e-02})} &
\textcolor{black}{1.19e-01(1.96e-01)} &
\textcolor{black}{5.95e-02(8.82e-02)} &
\textcolor{black}{5.45e-02(1.08e-01)} &
\textcolor{black}{7.96e-02(1.22e-01)}
 \\
    \bottomrule
    \end{tabular}}
\end{table*}

\begin{figure}[t]
	\centering
	\subfloat[Flickr]{
		\includegraphics[width=0.45\linewidth]{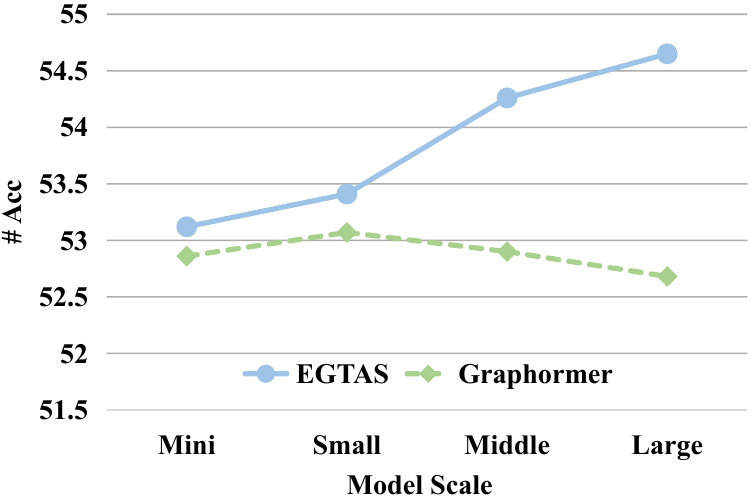}}
	\quad
    \subfloat[ogbn-arxiv]{
		\includegraphics[width=0.45\linewidth]{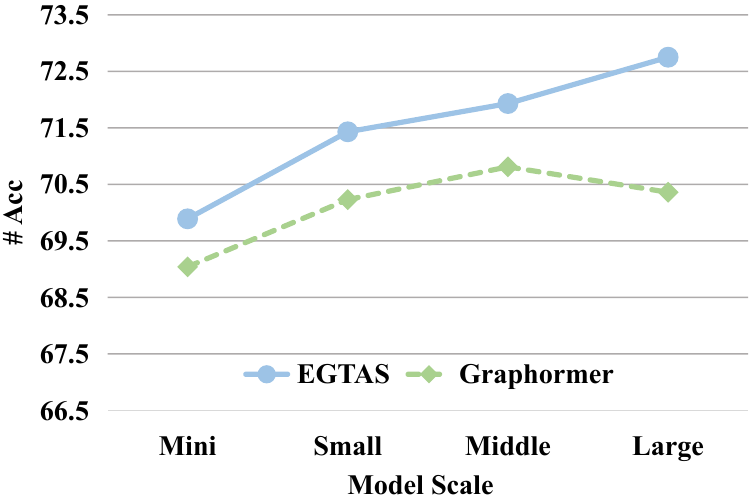}}
	\quad	
    \caption{\textcolor{black}{Accuracy comparison of EGTAS and Graphormer at different model scales on Flickr and ogbn-arxiv datasets.}}
	\label{AML}
\end{figure}

This section further evaluates the effectiveness of different surrogate models. We uniformly sample 50 architectures and train them on all datasets for 1e+4 maximum steps. 90\% of the samples are randomly selected to fit the surrogate model and the remaining 10\% are used for testing. Two popular evaluation metrics, MSE and KTau, are adopted. Table \ref{SM} lists the comparison of six classic surrogate models on all datasets for the node classification task over ten runs. From Table \ref{SM}, no single surrogate model outperforms all others in all cases. Therefore, in EGTAS, the best-performing surrogate model is embedded into the evolutionary search.

\textcolor{black}{We use the Flickr and ogbn-arxiv datasets as examples to illustrate the impact of a critical parameter, model scale, on the performance of EGTAS.} Specifically, EGTAS explores the proposed search space at different model scales. Fig. \ref{AML} shows the accuracy of EGTAS and manually designed architectures Graphormer at different model scales. We can observe that the accuracy of EGTAS increases as the model scale increases. Graphormer that simply stacks Transformer blocks may suffer from performance degradation. This phenomenon further illustrates that searching topologies and graph-aware strategies to extract deep features can alleviate the over-smoothing problem.

\section{Conclusion}
This paper introduces the EGTAS framework for a comprehensive exploration of topologies and graph-aware strategies in graph Transformers. Specifically, we design an extensive search space at the macro and micro levels, which covers existing state-of-the-art graph Transformers. EGTAS presents a surrogate-assisted evolutionary search to find high-performance architectures automatically. A generic encoding-based surrogate model is applied to reduce the model evaluation cost, thus promoting the search efficiency of EGTAS. \textcolor{black}{Extensive experiments on benchmark and real-world graph datasets demonstrate that EGTAS outperforms existing hand-designed and GNAS methods. In real-world scenarios, EGTAS can find multiple graph Transformers with varying scales and performances for downstream tasks in a single execution. Such flexibility ensures that users can autonomously select the most suitable architecture based on their deployment requirements.}

In the future, we will focus on multiple objectives, such as the number of parameters and latency, to recommend practical graph Transformers for different computing scenarios. Due to the complexity of graph-structured data, Transformer architecture search on other graph tasks is also a promising direction, such as link prediction and recommendation \cite{LI2023109874}. \textcolor{black}{In addition, graph learning with heterophily \cite{zheng2022graph} or homophily \cite{ojha2024affinitybased} is receiving increasing attention. The exploration of Transformer architectures tailored to special graphs merits deeper investigation.} As is often the case with GNAS, surrogate model construction on large-scale graphs suffers from computational bottlenecks. Evaluating multiple sampling architectures in parallel offers an efficient solution. Implementing the parallelization of EGTAS remains a topic for future work.

\bibliographystyle{IEEEtran}
\bibliography{my}

\newpage

\begin{IEEEbiography}[{\includegraphics[width=1in,height=1.25in,clip,keepaspectratio]{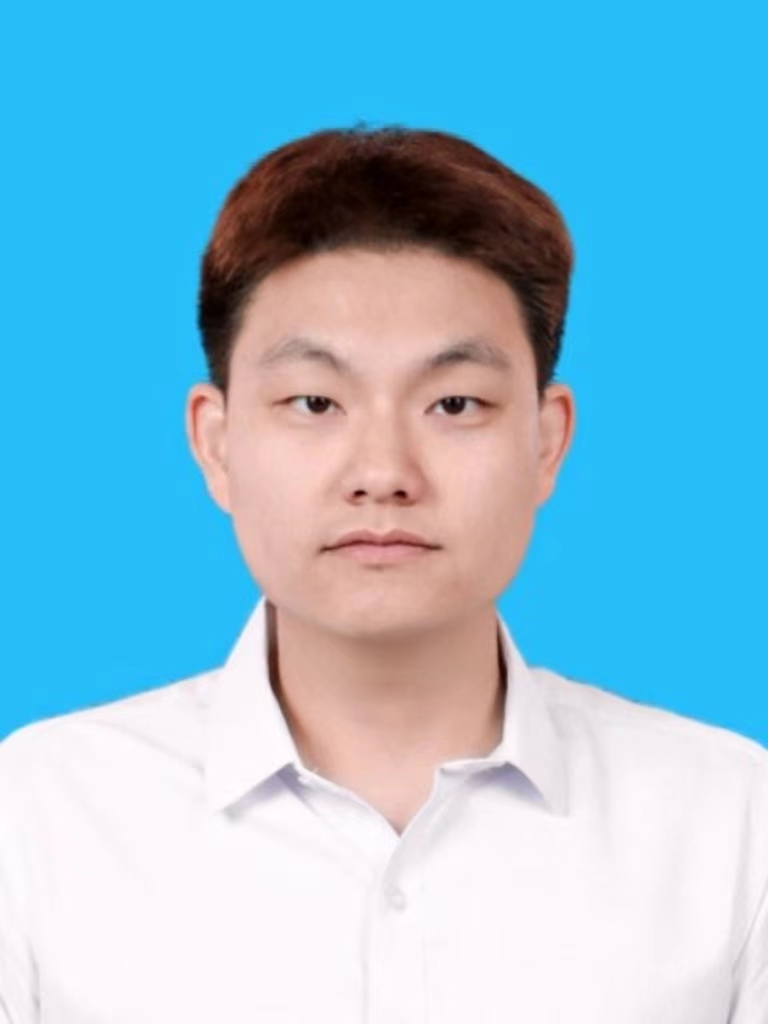}}]{Chao Wang} received the B.S. degree in
intelligent science and technology in 2019 from Xidian University, Xi’an, China, where he is currently pursuing the Ph.D. degree with the Key Laboratory of Intelligent Perception and Image Understanding of Ministry of Education, School of Artificial Intelligence Xidian University, Xi’an, China. 

His current research interests include multi-task learning and optimization, evolutionary computation, and graph learning.
\end{IEEEbiography}

\begin{IEEEbiography}[{\includegraphics[width=1in,height=1.25in,clip,keepaspectratio]{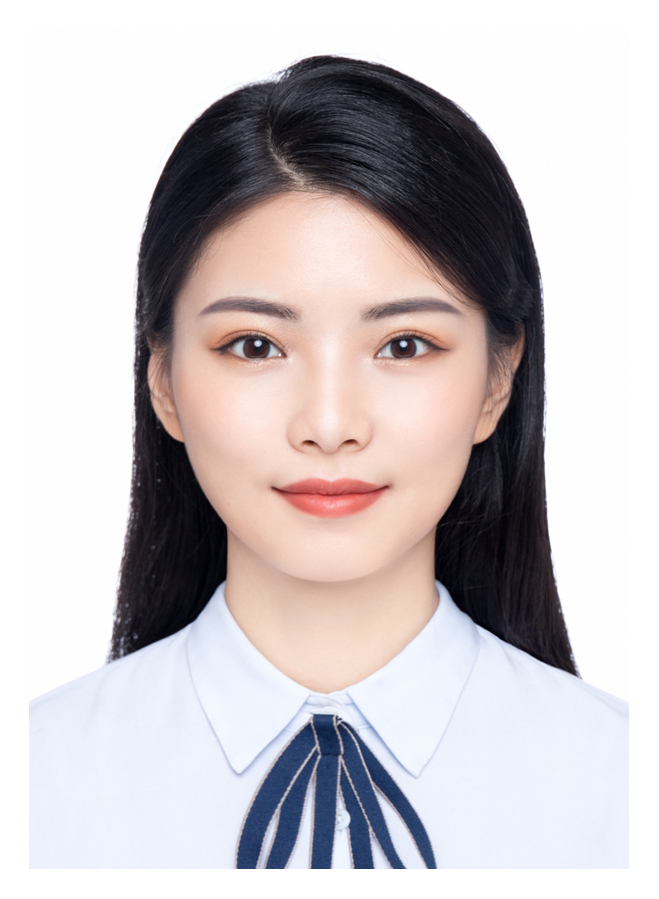}}]{Jiaxuan Zhao} (Student Member, IEEE) received the B.S. degree in materials science and engineering from Xidian University, Xi’an, China, in 2019. She is currently pursuing the Ph.D. degree with the Key Laboratory of Intelligent Perception and Image Understanding of Ministry of Education, School of Artificial Intelligence Xidian University, Xi’an, China. 

Her research interests include multimodal fusion, evolutionary computing, and image understanding.
\end{IEEEbiography}

\begin{IEEEbiography}[{\includegraphics[width=1in,height=1.25in,clip,keepaspectratio]{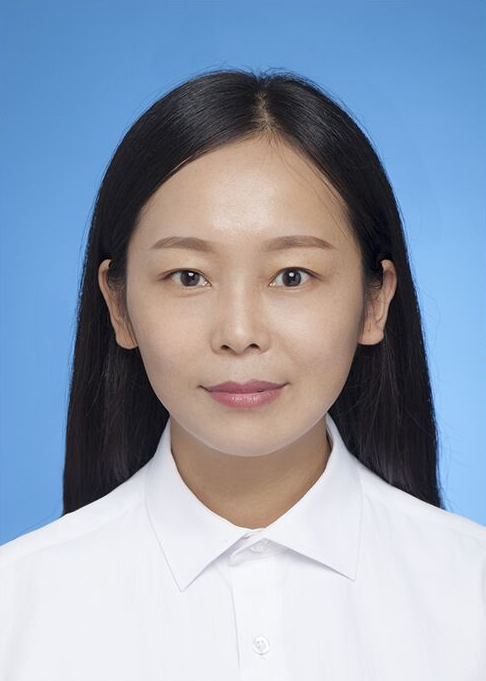}}]{Lingling Li}
(Senior Member, IEEE) received the B.S. and Ph.D. degrees from Xidian University, Xi’an, China, in 2011 and 2017, respectively.

From 2013 to 2014, she was an exchange Ph.D. student with the Intelligent Systems Group, Department of Computer Science and Artificial Intelligence, University of the Basque Country UPV/EHU, Spain. She is an associate Professor with the Key Laboratory of Intelligent Perception and Image Understanding, Ministry of Education, School of Artificial Intelligence, Xidian University. Her research interests include image processing, deep learning, and pattern recognition.
\end{IEEEbiography}

\begin{IEEEbiography}[{\includegraphics[width=1in,height=1.25in,clip,keepaspectratio]{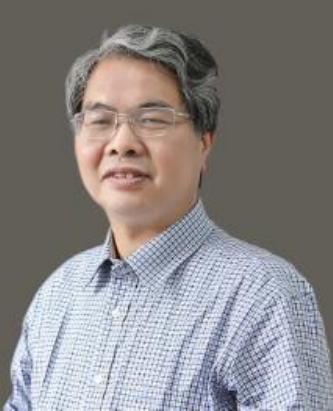}}]
{Licheng Jiao} (Fellow, IEEE) received his B.S. degree from Shanghai Jiaotong University, Shanghai, China, in 1982, and his M.S. and PhD degrees from Xi’an Jiaotong University, Xi’an, China, in 1984 and 1990, respectively.

Since 1992, he has been a professor at Xidian University. Now, he is a distinguished professor and serves in the School of Artificial Intelligence, Xidian University, Xi’an. He currently serves as the Director of the Key Laboratory of Intelligent Perception and Image Understanding, which is affiliated with the Ministry of Education of China. He has been a member of the Academia Europaea. His research interests include machine learning, deep learning, natural computation, remote sensing, image processing, and intelligent information processing.

Prof. Jiao is the chairperson of the Awards and Recognition Committee in IEEE Xi'an Branch, the 6th and 7th Vice Chairmen of the Chinese Association of Artificial Intelligence, the chairperson of the Asian Society for Computational Intelligence, the Fellow of IEEE/IET/CAAI/CIE/CCF/CAA/CSIG/AAIA/AIIA/ACIS, a councilor of the Chinese Institute of Electronics, a committee member of the Chinese Committee of Neural Networks, an expert of the Academic Degrees Committee of the State Council, and an ESI highly cited scientist.
\end{IEEEbiography}

\begin{IEEEbiography}[{\includegraphics[width=1in,height=1.25in,clip,keepaspectratio]{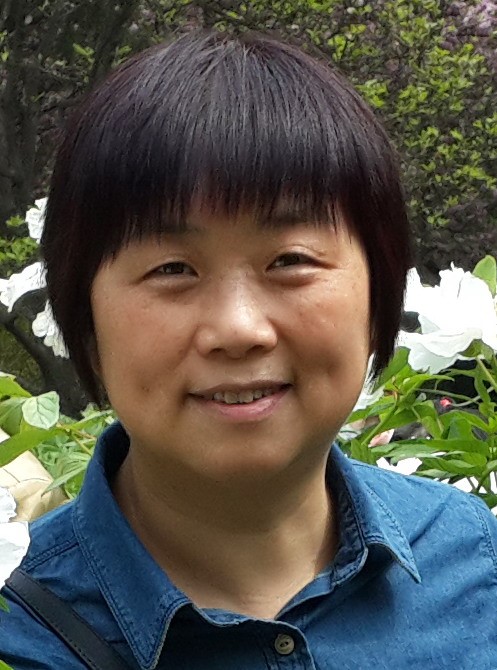}}]{Fang Liu}
(Senior Member, IEEE) received the B.S. degree in computer science and technology from the Xi’an Jiaotong University, Xi’an, China, in 1984 and the M.S. degree in computer science and technology from the Xidian University, Xi’an, in 1995. 

She is currently a Professor with the School
of Artificial Intelligence, Xidian University. Her
research interests include signal and image processing, synthetic aperture radar image processing,
multi-scale geometry analysis, learning theory and
algorithms, optimization problems, and data mining.
\end{IEEEbiography}

\begin{IEEEbiography}[{\includegraphics[width=1in,height=1.25in,clip,keepaspectratio]{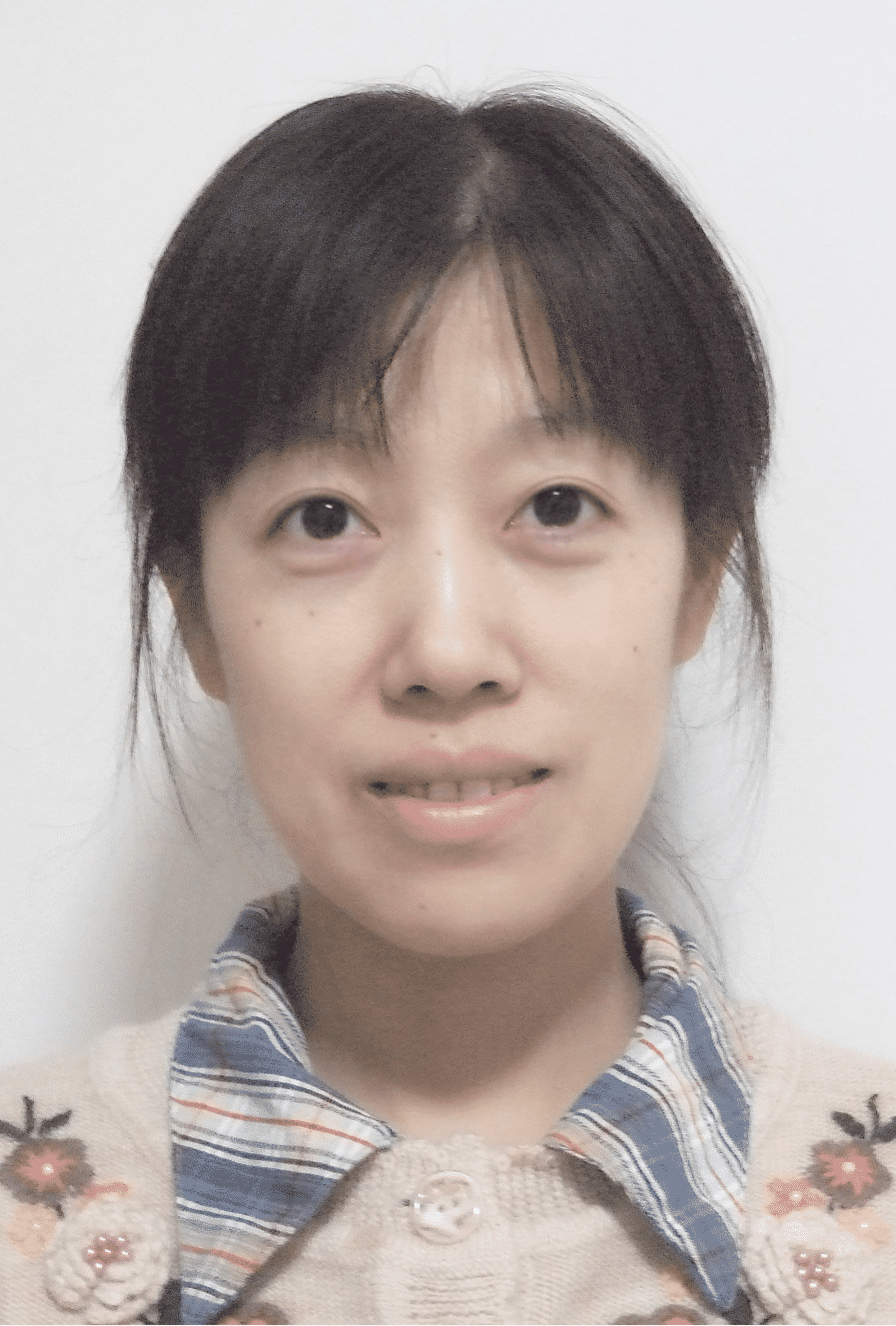}}]{Shuyuan Yang}
(Senior Member, IEEE) received the
B.A. degree in electrical engineering, and the M.S.
and Ph.D. degrees in circuit and system from Xidian
University, Xi’an, China, in 2000, 2003, and 2005,
respectively.

She has been a Professor with the School
of Artificial Intelligence, Xidian University. Her
research interests include machine learning and
multiscale geometric analysis.
\end{IEEEbiography}
\enlargethispage{-15.2cm}

\end{document}